\documentclass{article} % For LaTeX2e
\usepackage{iclr2026_conference,times}
\usepackage{graphicx}
 \usepackage{multirow}
 \usepackage{amsthm}
\usepackage{algorithm2e}
\SetKwInOut{Input}{Input}
\SetKwInOut{Output}{Output}
\usepackage{amsmath,amsfonts,bm}

\def\eqref#1{equation~\ref{#1}}
\def\1{\bm{1}}

\DeclareMathAlphabet{\mathsfit}{\encodingdefault}{\sfdefault}{m}{sl}
\SetMathAlphabet{\mathsfit}{bold}{\encodingdefault}{\sfdefault}{bx}{n}

\DeclareMathOperator*{\argmin}{arg\,min}

\usepackage{hyperref}
\usepackage{url}
\usepackage{cleveref}
\usepackage{booktabs}
\usepackage{caption}
\usepackage{xspace}
\def\onedot{.\xspace}
\def\@onedot{\ifx\@let@token.\else.\null\fi\xspace}
\def\eg{\emph{e.g}\onedot} \def\Eg{\emph{E.g}\onedot}
\def\ie{\emph{i.e}\onedot} \def\Ie{\emph{I.e}\onedot}

\makeatother

\usepackage{subcaption}
\usepackage{wrapfig}
\usepackage{enumitem}

\newcommand{\task}{Noise-Aware Generalization}
\newcommand{\taskspace}{Noise-Aware Generalization }
\newcommand{\taskAbbr}{NAG}
\newcommand{\taskAbbrspace}{NAG }

\newcommand{\methodspace}{Domain Labels for Noise Detection }
\newcommand{\methodAbbr}{DL4ND}
\newcommand{\methodAbbrspace}{DL4ND }

\title{\task: Robustness to In-Domain Noise and Out-of-Domain Generalization}

\author{Siqi Wang, Aoming Liu, Bryan A. Plummer \\
Department of Computer Science\\
Boston University\\
\texttt{\{siqiwang,amliu,bplum\}@bu.edu} 
}

\iclrfinalcopy % Uncomment for camera-ready version, but NOT for submission.
\begin{document}

\maketitle
\begin{abstract}

Methods addressing Learning with Noisy Labels (LNL) and multi-source Domain Generalization (DG)  use training techniques to improve downstream task performance in the presence of label noise or domain shifts, respectively.  Prior work often explores these tasks in isolation, and the limited work that does investigate their intersection, which we refer to as \taskspace (\taskAbbr), only benchmarks existing methods without also proposing an approach to reduce its effect. We find that this is likely due, in part, to the new challenges that arise when exploring \taskAbbr{}, which does not appear in LNL or DG alone. For example, we show that the effectiveness of DG methods is compromised in the presence of label noise, making them largely ineffective.  Similarly, LNL methods often overfit to easy-to-learn domains as they confuse domain shifts for label noise.  Instead, we propose \methodspace (\methodAbbr), the first direct method developed for \taskAbbr{} which uses our observation that noisy samples that may appear indistinguishable within a single domain often show greater variation when compared across domains.  We find \methodAbbr{} outperforms DG and LNL methods, including their combinations, even when simplifying the \taskAbbr{} challenge by using domain labels to isolate domain shifts from noise.  Performance gains up to 12.5\% over seven diverse datasets with three noise types demonstrates \methodAbbr's ability to generalize to a wide variety of settings\footnote{Code: \url{https://github.com/SunnySiqi/Noise-Aware-Generalization}}.
\end{abstract}    
\section{Introduction}

\begin{figure}[t]
  \centering
    \includegraphics[width=.85\textwidth]{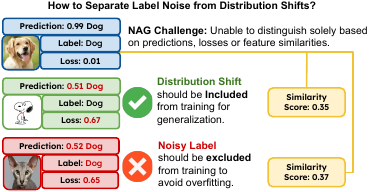}
    %\vspace{-2mm}
    \caption{A key challenge in \taskAbbrspace is being able to separate distribution shifts due to domain from shifts due to noise.  However, in practice these samples can be hard to distinguish between each other, which makes mitigating the effect of label noise challenging as many LNL methods require some mechanism of detecting the noise (\eg,~\citet{li2023disc, karim2022unicon, zhao2024estimating}).  This represents a new challenge that only emerges when considering learning with noisy labels and domain generalization together, highlighting the need to explore \taskAbbr.}
  \label{fig:intro}
  %\vspace{-2mm}
\end{figure}

Domain Generalization (DG) methods train models to generalize to unseen target domains by learning from multiple source domains (\eg,~\citet{cha2022miro, cha2021swad, wang2023sharpness, arjovsky2019invariant, kamath2021does, chen2022pareto, chen2024understanding, rame2022fishr, lin2022zin}). However, these methods tend to ignore label noise, which appears naturally in many datasets (\eg,~\citet{chen2024chammi,clothing1m,li2017webvisiondatabasevisuallearning,song2019selfie}), including those used in DG benchmarks~\citep{teterwak2024large}.  Despite this, limited prior work has explored the effect of label noise in DG settings~\citep{qiao2024understanding, seo2020learning}, which simply evaluates existing methods without proposing solutions to improve robustness.  A naive approach to mitigate label noise issues  would be simply to combine DG methods with those from the Learning with Noisy Labels (LNL) literature (\eg,~\citet{natarajan2013LNL, arpit2017closer, song2022survey, xia2021sample, xia2023combating, wei2022self, liu2021noise, song2024no, cordeiro2023longremix, shen2019learning_iterative, WangLNLK2024}).  Generally speaking, LNL methods work by detecting potential noise and then either removing~\citep{crust, kim2021fine}, relabeling~\citep{karim2022unicon, li2023disc}, or downweighting~\citep{liu2020early, liu2023identifiability} these samples.  However, as shown in \Cref{fig:intro}, separating distribution shifts due to noise from distribution shifts due to domain is challenging since they look similar according to feature similarity or analyzing loss values (an observation echoed in related work on out-of-distribution detection~\citep{humblot2024noisy}).

\begin{figure}[t]
  \centering
  \begin{minipage}{0.55\textwidth}
    \includegraphics[width=\textwidth,trim=0cm 5.8cm 12.3cm 0cm,clip]{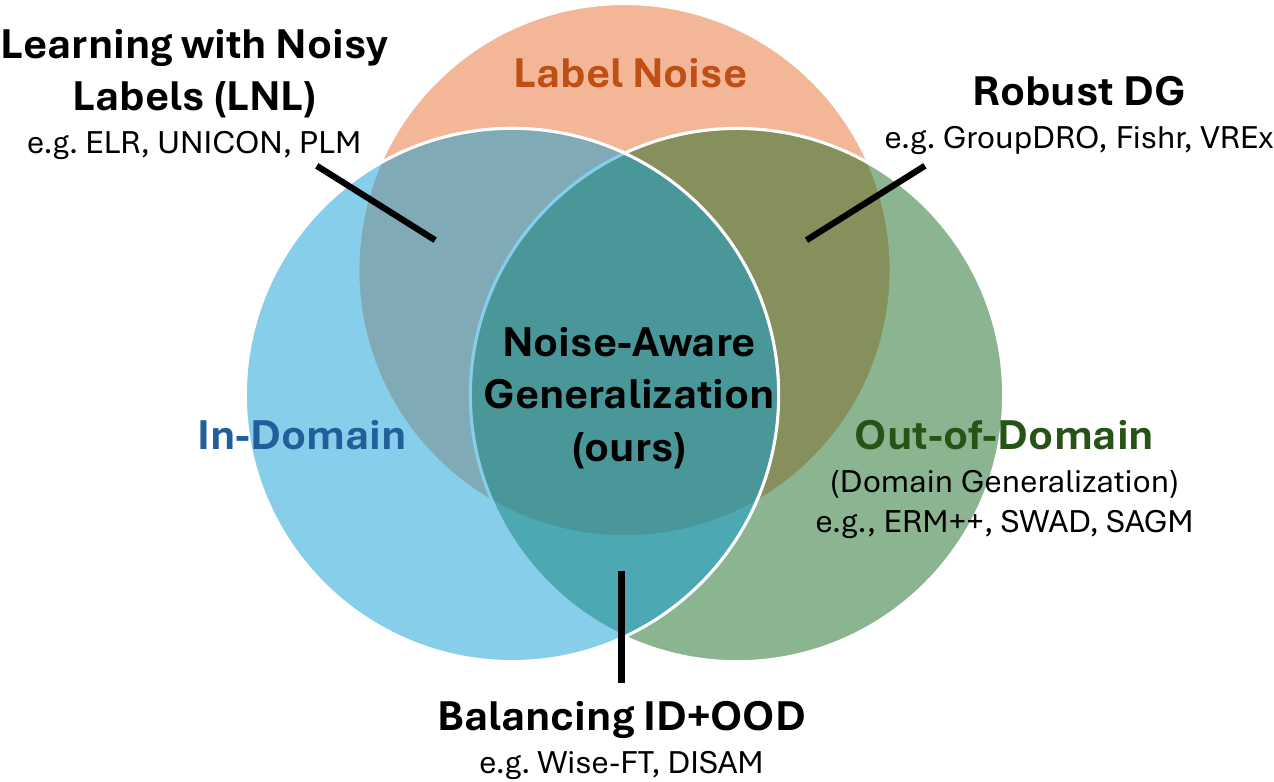}
  \end{minipage}%
  \hfill
  \begin{minipage}{0.4\textwidth}
    \captionof{figure}{DG methods often ignore in-domain performance (\eg~\citet{sagawa2019distributionally, rame2022fishr, krueger2021out,zhang2024domain, wortsman2022robust}), label noise (\eg~\citet{teterwak2023erm++, cha2022miro, cha2021swad, wang2023sharpness}), or both. LNL methods handle in-domain label noise, but not domain shifts~\citep{liu2020early, li2023disc, karim2022unicon, zhao2024estimating}. In contrast, \taskAbbrspace imitates real-world applications by requiring methods that do well on all data when trained on noisy data.}
    \label{fig:related_work}
  \end{minipage}
 % \vspace{-2mm}
\end{figure}

We refer to the intersection of DG and LNL as \task{} (\taskAbbr), where the goal is to maximize both in-domain (ID) and out-of-domain (OOD) performance when training on noisy, multi-domain datasets.  This makes \taskAbbr{} methods more practical as they can be applied to a wider range of settings than DG or LNL alone.  In particular, there are three major components in \Cref{fig:related_work} we use to compare \taskAbbr{} to related tasks: 1) methods that are evaluated on ID performance,  2) methods that are evaluated on OOD performance, 3) methods that are evaluated when trained with label noise.  The benchmarks used by method in prior work would cover two of these settings at most.  For example, DG methods either ignore in-domain performance (\eg,~\citet{sagawa2019distributionally, rame2022fishr, krueger2021out}) or label noise (\eg,~\citet{teterwak2023erm++, cha2022miro, cha2021swad, wang2023sharpness}). LNL methods handle noise, but do not consider domain shifts~\citep{liu2020early, li2023disc, karim2022unicon, zhao2024estimating}.  However,  it is common to deploy a model that should perform well on both ID and OOD data, and label noise is expected as it is often prohibitively expensive to collect perfectly clean data (\eg, where expertise is required or annotators simply disagree~\citep{chen2024chammi,tanno2019learning,vorontsov2021label}).

We begin by exploring \taskAbbrspace by analyzing the challenges through experiments on a synthetic noise dataset, providing foundational insight for further research.  In particular, we highlight the issues with developing LNL-style methods that aim to detect noise, partly highlighted in \Cref{fig:intro}, where identifying that distribution shifts in the training set are due to noise rather than domain shifts is challenging.  A naive solution to this problem would be to separate samples into their respective domains.  In effect, this would theoretically reduce the problem to the traditional LNL task, where a method would identify noisy samples within a single domain. However, spurious features in some domains make them more challenging to detect.  

\begin{figure}
    \centering
    \includegraphics[width=.85\linewidth,trim=0cm 7.5cm 7.8cm 0cm,clip]{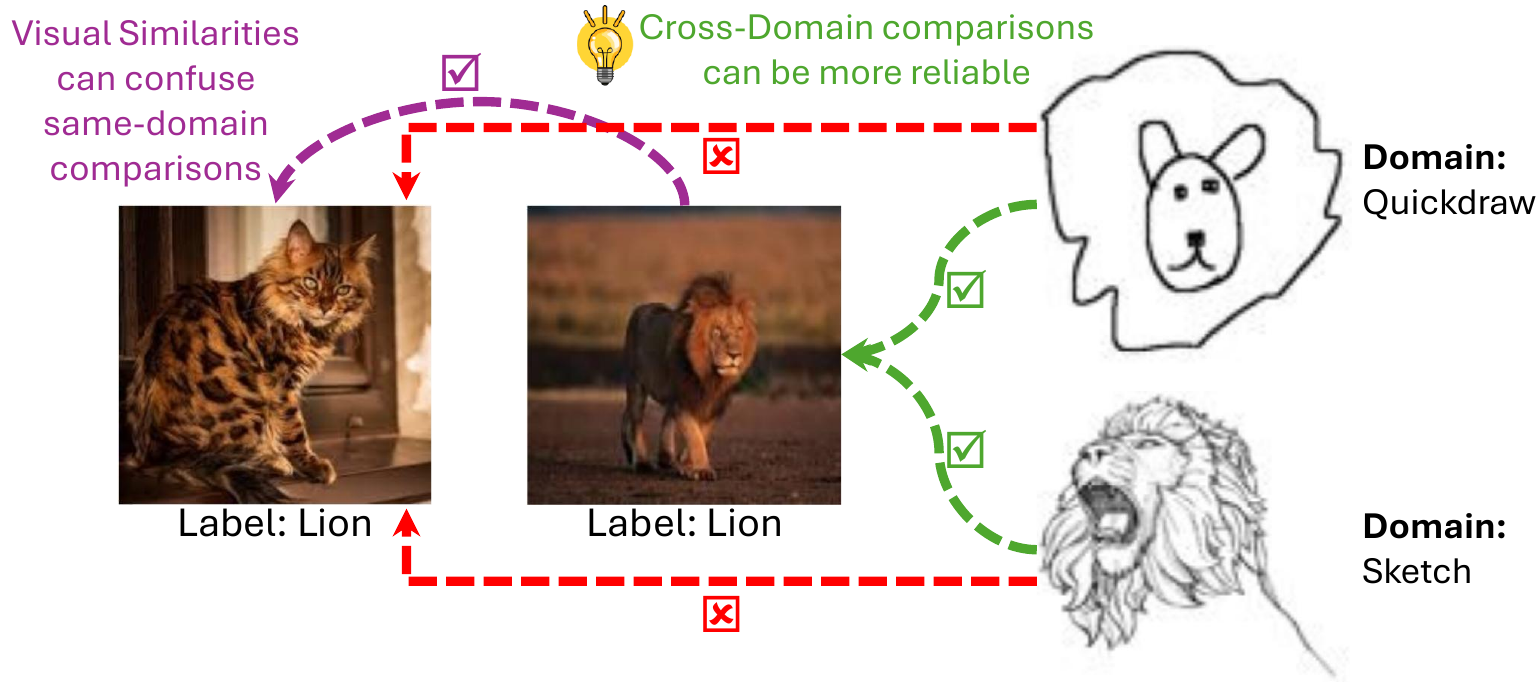}
    \vspace{-2mm}
    \caption{\textbf{Images in example embedding space.} Comparing images within the same domain (shown in purple) can rely on spurious features, such as similar colors, resulting in labels being incorrectly identified as clean.  However, cross-domain comparisons (shown in red and green) must rely on intrinsic features, thereby providing a more accurate estimate of an image's label.}
    \label{fig:lion}
    %\vspace{-2mm}
\end{figure}

For example, the two images labeled as``lion" in \Cref{fig:lion} have similar colors, which a model may inadvertently use as most lions also have the same colors.  Thus, separating them within the photo domain would be harder than across domains where they must rely on intrinsic, domain-agnostic features. This observation motivates our proposed method, \methodspace (\methodAbbr), which identifies noisy samples by extracting (class, domain) proxies from low-loss samples. Then we use these proxies for reliable noise detection through cross-domain comparisons. 
We show that  \methodAbbrspace can improve performance over other LNL methods on its own, or can be combined with DG methods for further gains.
Experiments with 12 state-of-the-art DG and LNL methods, along with 20 combination methods on three real and four synthesized noise datasets, show that \methodAbbrspace performs up to 12.5\% better than prior work.

Our contributions are summarized below.
\vspace{-2mm}
\begin{itemize}[nosep,leftmargin=*]
    \item We highlight the challenges of real-world datasets that exhibit both label noise and domain shifts in diverse fields, including web data~\citep{fang2013unbiased} and biological imaging~\citep{chen2024chammi}. 
    \item We investigate the underexplored setting we call \task{} (\taskAbbr), which focuses on training a robust network under ID noise while ensuring good generalization to OOD data. We analyze the limitations of existing approaches and their naive combinations to provide insight into \taskAbbr{} and its challenges.
    \item We propose \methodAbbr, a novel noise detection method that can be used on its own or in combination with prior work in DG, showing a promising solution to \taskAbbr.
\end{itemize}

% merge this section into Intro + Method
\section{Related Work}

As discussed in the Introduction and summarized in \Cref{fig:related_work}, \taskAbbrspace combines elements from both the Domain Generalization (DG) and Learning with Noisy Labels (LNL), producing new challenges not present when exploring each task alone.  DG methods (\eg,~\citet{gulrajanisearch,Li2017DeeperBA,Li2017LearningTG, Li2019EpisodicTF,Li2018DomainGW,Li2018DeepDG, Muandet2013DomainGV,teterwak2023erm++}) often focus on learning domain-invariant features (\ie, aligning their representations).  In contrast, LNL methods typically aim to minimize the effect of label noise by reweighting samples based on an estimate of cleanliness (\eg,~\citet{scott2015rate, liu2015classification, menon2015learning, patrini2017making, li2021provably, zhang2021learning, kye2022learning, cheng2022instance, liu2023identifiability, li2022estimating, vapnik2013constructive, yong2022holistic}) or detecting and removing (or relabeling) the noisy samples (\eg,~\citet{hu2021p, torkzadehmahani2022confidence, nguyen2019self, tanaka2018joint,Hou_2025_CVPR, li2022selective, feng2021ssr,li2023disc,song2024no,xia2023combating,cordeiro2023longremix,wei2022self,xia2021sample}).   However, DG methods often ignore the effect of label noise and only evaluate on out-of-domain (OOD) generalization.  In contrast, LNL methods study the effect of label noise on in-domain (ID) data, but ignore domain shifts.  Thus, these methods may not generalize in real-world settings like explored by \taskAbbr, where (at least some) label noise is expected and models have to perform well on ID and OOD settings.

While there is limited DG work that evaluates based on ID and OOD performance (\eg,~\citet{zhang2024domain,wortsman2022robust}), these methods do not explore the effect of label noise.  Similarly, prior work that evaluates existing DG method's performance with label noise do not develop new techniques for this setting~\citep{qiao2024understanding, seo2020learning}, only exploring preexisting approaches.  Some work that explores the intersection of Domain Adaptation (DA) and LNL proposes custom methods (\eg,~\citet{10.1609/aaai.v33i01.33014951,9921307,9488210,10.1109/TMM.2022.3205457,10.1145/3581783.3612296,Yin_Feng_Yan_Song_Peng_Wang_2025}), but as DA is typically performed a single source domain, they do not have to learn to separate domain shifts from label noise (\ie, they do not have to address the challenge in \Cref{fig:intro}).  As such, many of the DA+LNL methods closely match those in the LNL literature (see Appendix~\ref{sec:dalnl} for examples), with similar limitations. Additionally, prior work in DG+LNL and DA+LNL are typically not evaluated on their in-domain performance. As our experiments show, these limitations mean that they do not generalize well to \taskAbbr, often performing on par with simple ERM~\citep{gulrajani2020search-erm}. Thus, exploring \taskAbbrspace is a necessary step to creating methods that generalize to real-world settings.

\section{\task{} (\taskAbbr)}
\label{sec:nag}

Consider a multi-domain dataset $\mathcal{D}$ with $m$ source domains:  $\mathcal{D} = \left\{ \mathcal{D}_1, \mathcal{D}_2, \ldots, \mathcal{D}_m \right\}$, where each $\mathcal{D}_i = \left\{(x_{i,j}, \tilde{y}_{i,j})\right\}_{j=1}^{n_i}$ represents samples from domain $i$ with $x_{i,j}$ as the input and $\tilde{y}_{i,j}$ as the label, potentially noisy and the true label $y_{i,j}$ is unknown. The goal in \taskAbbrspace is to learn a featurizer $f_\theta(\cdot)$ parameterized by $\theta$ that performs well in all source domains $\{\mathcal{D}_i\}_{i=1}^m$ and generalizes to unseen domain(s) $\mathcal{D}_{target}$, despite the presence of label noise. For convenience in describing the equation in the rest of the section, we denote the domain of an input $x$ as $D(x)$ and its class label as $Y(x)$. We use $d(\cdot)$ to represent the cosine distance between feature embeddings.

To help better understand \taskAbbr, we perform a preliminary analysis using RotatedMNIST~\citep{ghifary2015domain}, chosen for its simplicity and clear feature structure.  Note that we also validate these observations on more complex datasets in our experiments. We select four rotation angles (0°, 15°, 30°, and 45°) to represent different domains. Pairwise noise is introduced by randomly flipping 30\% of the labels between four confusing digit pairs: (0, 6), (1, 7), (3, 5), and (4, 9). Our experiments use a ResNet50~\citep{he2016deep} trained via ERM~\citep{gulrajani2020search-erm}.

\subsection{Can we separate class shifts from domain shifts?}
\label{sec:low_loss}

Detecting samples that may stem from label noise is a key capability for LNL-style methods.  However, as \taskAbbrspace also requires generalizing across domains, an important capability that distinguishes \taskAbbrspace from related work is being able to separate samples with shifts due to domain from shifts due to label noise during training (partially highlighted in \Cref{fig:intro}).  However, measuring distributional differences between samples (\eg, by computing feature distances between all sample pairs) can be computationally expensive.  Thus, LNL methods often rely on having some canonical representation of a class to measure similarity of a particular sample (\eg,~\citet{crust,song2024no,zhao2024estimating}), for example, by averaging the features of samples of that class.  If shifts in domain were closer to this canonical representation than shifts in category, then we could successfully identify potential noise.  

More formally, continuing the earlier notation, let us group samples in domain-class sets, \ie,
\begin{equation}
    G_{c,i} = \{ x_j \mid Y(x_j) = c, \mathcal{D}(x_j) = i \}.
\label{notation:group}
\end{equation}
Let $\bar{g}_{c,i} = \frac{1}{|G_{c,i}|}\sum f_\theta(G_{c,i})$ denote the average of the set of learned features for group $G_{c,i}$. We assume the existence of a featurizer $f_\theta(\cdot)$ such that:  
\begin{equation}
    \text{d}(f_\theta(G_{c,\hat{i}}), \bar{g}_{c,i}) < \text{d}(f_\theta(G_{\hat{c},i}), \bar{g}_{c,i}), \text{ where } i\neq\hat{i} \text{ and } c\neq\hat{c}.
    \label{eq:assumption1}
\end{equation}
This suggests that we can perfectly separate class shifts from domain shifts if we can train a featurizer such that, for each sample, the distance to other samples within the same class (across different domains) is smaller than the distance to samples within the same domain (but different classes).  

\begin{figure}[t]
\centering
%\begin{minipage}[c]{0.56\textwidth}
 \includegraphics[width=.8\linewidth]{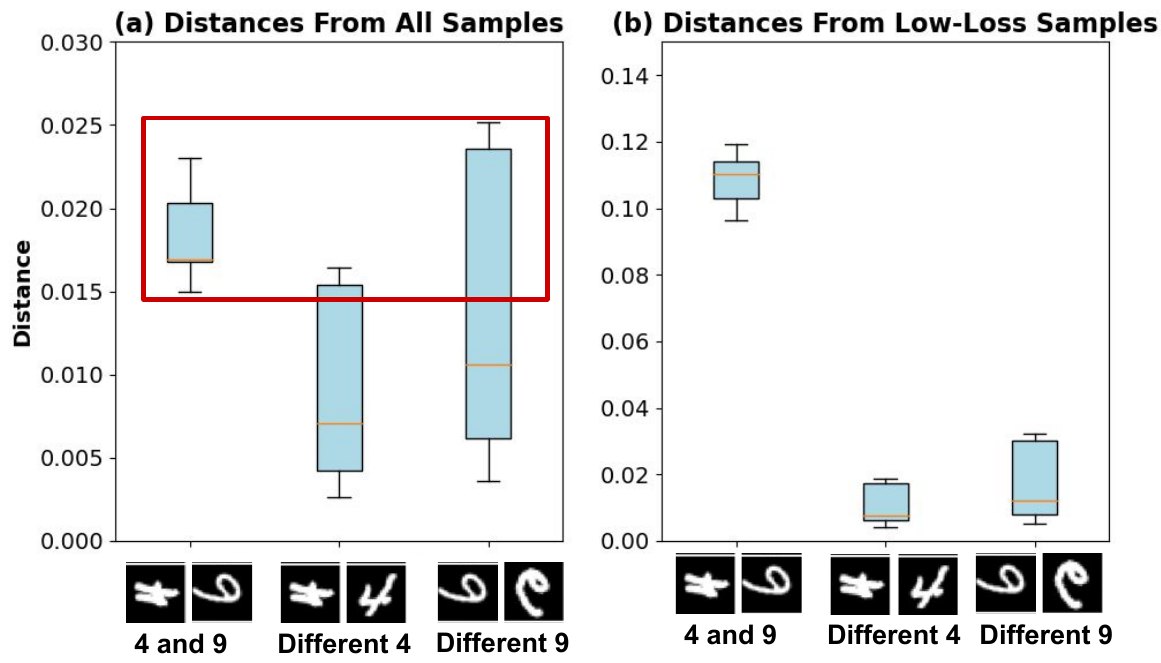}
%\vspace{10pt}
%\vspace{-2mm}
%\end{minipage}%
%\hspace{0.3cm}
%\hfill
%\begin{minipage}[c]{0.4\textwidth}
\caption{\textbf{Box plot of distance distributions between samples and their canonical class representation $\bar{g}_{c,i}$.} (a) Reports distances between training samples and an $\bar{f_D}$ computed by averaging all training samples in its class, domain group $G_{c,i}$. The red box highlights overlapping distributions, indicating the challenge of distinguishing samples with class and domain shift. (b) $\bar{g}_{c,i}$ is calculated from low-loss samples which shows shifts in class can be separated from shifts in domain. See \Cref{sec:low_loss} for additional discussion.}
  \label{fig:toy}
%\end{minipage}
%\vspace{-2mm}
\end{figure}

\Cref{fig:toy} explores this possibility by showing the distribution of distances in the presence of both shifts in domain and shifts in class on RotatedMNIST~\citep{ghifary2015domain}.  As seen in the red box in \Cref{fig:toy}(a), where $\bar{g}_{c,i}$ is computed over all samples in the group $G_{c,i}$, there is a significant portion of the shifts in category and domain that cannot be distinguished from each other (verifying the example in \Cref{fig:intro} occurs in practice).  Thus, \Cref{eq:assumption1} is not satisfied and we cannot use this approach to identify noisy samples.  However, as shown in \Cref{fig:toy}(b), we find that we can satisfy \Cref{eq:assumption1} by creating $\bar{g}_{c,i}$ with samples we have high confidence are not noise. For example, by using only low-loss samples early in training (\ie, before the model begins overfitting)that prior work has shown are typically clean~\citep{liu2020early,choi2025eldet}. See further validation of these observations on more complex datasets in \Cref{sec:exp}.

\subsection{Is it really that important to satisfy \Cref{eq:assumption1}?}

As discussed in the previous section, if we are not careful about how we measure feature similarity between samples then we can end up in a situation like for samples in red box in \Cref{fig:toy}(a), where we simply can't distinguish between shifts due to class and shifts due to domain. However, one could argue that if we are unsure about these samples, we should simply mitigate their contribution to model training (\eg, by downweighting their loss or removing them entirely).  \Ie, if we can still learn the decision boundary between classes without them, then these samples are unnecessary.

To quantify sample importance to the final decision boundary, we trained an SVM on RotatedMNIST and treat the resulting support vectors as the ``important" samples. We find that over 20\% of the support vectors fall within the red box region of \Cref{fig:toy}(a).  Thus, as we show in \Cref{sec:exp}, removing their contribution during training has a significant effect on the final decision boundary.

\section{\methodspace (\methodAbbr)}
\label{sec:DL4ND}

\begin{figure*}[t]
  \centering
    \includegraphics[width=\textwidth,trim=0cm 0cm 3.5cm 0cm,clip]{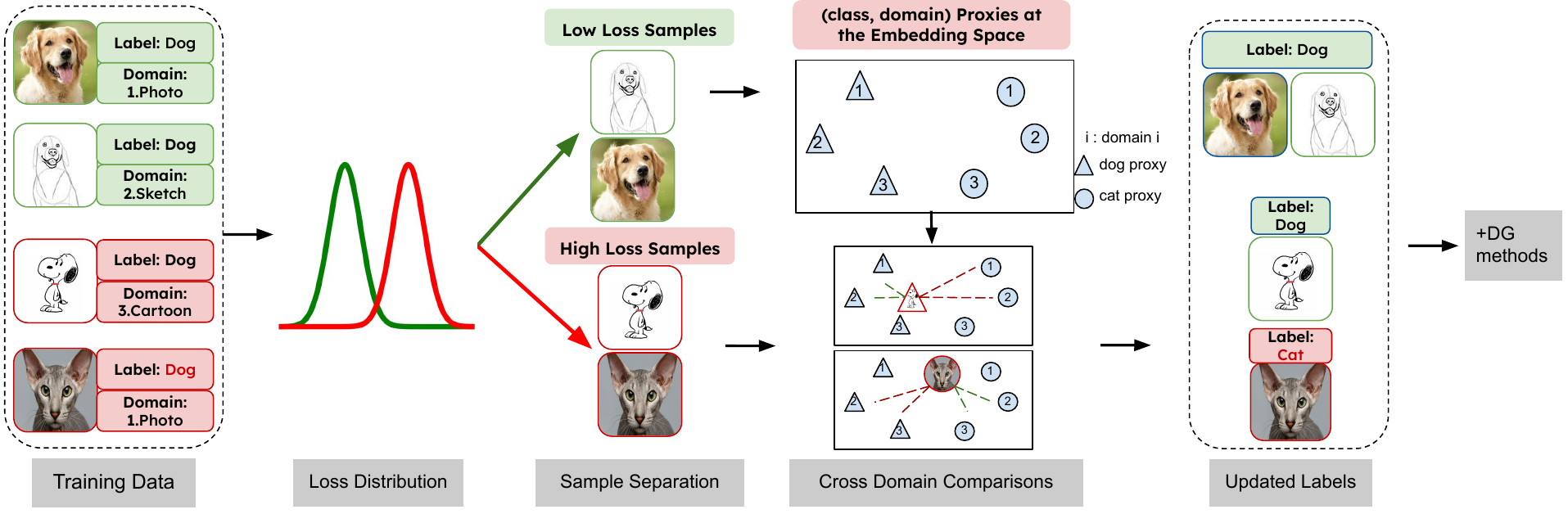}
    \vspace{-4mm}
    \caption{\textbf{\methodAbbrspace Framework}. After a brief warmup stage, the first step in \methodAbbrspace is to split samples into low-loss and high-loss groups using a Gaussian Mixture Model (GMM) based on the loss distribution. The low-loss sample's labels are frozen and used to generate (class, domain) proxies $\bar{g}_{c,i}$. High-loss samples are relabeled using the proxies with \Cref{eq:dland}. Then training resumes with the updated labels.  Note that in all stages, a DG method (\eg, ERM++~\citep{teterwak2023erm++}, SAGM~\citep{wang2023sharpness}, SWAD~\citep{cha2021swad}) can also be used in combination with \methodAbbr.  See \Cref{sec:DL4ND+DG_pipeline} for additional discussion.}
  \label{fig:pipline}
  %\vspace{-2mm}
\end{figure*}

In \Cref{sec:nag} we gave a formal definition of \taskAbbrspace and also provided an analysis that lead to a means of separating domain shifts from category shifts by using a small set of samples to create a proxy to represent the canonical features of a particular category.  If all datasets were as simple as RotatedMNIST, this might be sufficient to solve the \taskAbbrspace task.  However, in more complex datasets there can be many confounding factors, such as due to biases in a source domain (see \Cref{fig:lion}), that can confuse a model.  \Cref{sec:cross_domain} aims to reduce the impact of these spurious features via \methodAbbr, our proposed approach for detecting noise by using cross-domain comparisons.  \Cref{sec:DL4ND+DG_pipeline} describes how we fit our \methodAbbrspace noise detection method into our full \taskAbbrspace framework. See \Cref{fig:pipline} for an illustration summarizing our approach.

\subsection{Detecting Noise with Cross-Domain Comparisons}
\label{sec:cross_domain}

As illustrated in \Cref{fig:lion}, noisy samples may exhibit strong visual similarity to their incorrect noisy labels within a given domain.  This ``visual similarity" often arises from spurious features, such as background or color, which are domain-dependent and may not persist across different domains.  As we will discuss further below, these issues persist even if we were to simplify the \taskAbbrspace problem by using domain labels to keep training samples in their own domain, \ie, to perform single-domain comparisons to identify noise. These observations lead us to formulate a hypothesis: can cross-domain comparisons provide a stronger signal for a sample's true class than single-domain methods?

Let us perform a thought experiment with the example in \Cref{fig:lion}.  Since the lion images in the photo domain have a quintessential set of colors used for a typical lion, its reasonable that a model might use these correlations in the color space to identify if an image is a lion.  After all, if an image contained largely black pixels, then the model knows it is unlikely to be a lion in this domain.  However,  other domains may not have the same color bias, \eg, the sketch and draw images in \Cref{fig:lion}.  Thus, comparing images from these domains to the photos requires the model to rely on more lion's intrinsic features that generalize across domains.  Motivated by this observation (which we empirically validate in \Cref{sec:exp}), our \methodAbbrspace creates a new label $\hat{y_i}$ for sample $x_i$ identified as potential noise by finding the closest class representation from another domain $\bar{g}_{c,\hat{i}}$, \ie,:
\begin{equation}
    \hat{y_i} = \argmin_{\forall g_{c,\hat{i}}}  \text{d}(f_\theta(x_i), \bar{g}_{c,\hat{i}}), \text{ where } i\neq\hat{i}.
    \label{eq:dland}
\end{equation}
Note that this does not mean that $\hat{y_i} \neq y_i$ in all cases, as some $x_i$ may have their labels judged as correct according to \Cref{eq:dland}. Next we discuss how \Cref{eq:dland} is used in our full framework.

\subsection{\methodAbbrspace Framework}
\label{sec:DL4ND+DG_pipeline}

Prior work showed noisy labels have minimal effect on early stages of training~\citep{liu2020early}. In this stage the concept of a category is still being formed, often using general, easy-to-learn features.  Thus, LNL methods like \methodAbbrspace are typically applied in later stages where a model may start overfitting to noisy labels. This means we begin training using ERM~\citep{gulrajani2020search-erm}, or any other DG method, which is used in all stages of training.  After this warmup step, we first create our class proxies using the low-loss samples inspired by the discussion from \Cref{sec:low_loss}.

Specifically, rather than introducing a hyperparameter that manually controls the loss threshold, we assume the loss distribution is separated by a Gaussian Mixture Model (GMM) with two clusters. Samples belonging to the low-loss cluster serve as proxies, while high-loss samples require label updates through cross-domain comparisons. Low-loss samples are grouped by both domain and class (as shown in \Cref{notation:group}).  This means each (domain, class) pair has its own proxy representation, computed as the average feature of all low-loss samples in the same (domain, class) group. We assume these low-loss samples have clean labels and are kept frozen during training.

\begin{wraptable}{r}{0.48\textwidth} 
    \centering
    \vspace{-3mm}
    \setlength{\tabcolsep}{2pt}
    \caption{Top-1 accuracy on RotatedMNIST~\citep{ghifary2015domain} with 30\% asymmetric label noise. The baseline method creates $\bar{f_D}$ using all samples from its group $G_{c,i}$ with single-domain rather than cross-domain comparisons. See \Cref{sec:DL4ND+DG_pipeline} for discussion.}
    %\vspace{-2mm}
    \begin{tabular}{lccc}
        \toprule
        Method  & Label Acc. & ID  & OOD  \\
        \midrule
        Baseline & 75.7 & 87.7 & 87.9 \\
        \methodAbbr{} (ours)    & 98.1 & 98.1 & 97.8 \\
        \bottomrule
    \end{tabular}
    \label{tab:DL4ND_label_acc}
    %\vspace{-4mm}
\end{wraptable}

High-loss samples are relabeled using the cross-domain comparisons defined in \Cref{eq:dland}. Thus, \methodAbbrspace performs label refinement during training, requiring no additional data or learning overhead. This relabeling can be done periodically, as done in prior work~\cite{karim2022unicon,li2023disc}.   However, our experiments show that labeling only once can improve label quality. For example, \Cref{tab:DL4ND_label_acc} shows \methodAbbrspace on RotatedMNIST increased label accuracy from 75\% to 98\%, resulting in a 10\% boost in both ID and OOD accuracy. 

\section{Experiments}
\label{sec:exp}

\noindent\textbf{Datasets.} We use three real-world datasets (VLCS~\citep{fang2013unbiased}, CHAMMI-CP~\citep{chen2024chammi}, and PACS~\citep{Li2017DeeperBA}) and three additional synthetic noise datasets (OfficeHome~\citep{venkateswara2017deep}, TerraIncognita~\citep{beery2018recognition}, and DomainNet~\citep{peng2019moment}) to supplement our RotatedMNIST experiments reported earlier. \Cref{app:data_noise} discusses our noise types, which supplements the results using asymmetric noise on RotatedMNIST with real-world highly noisy dominant noise~\citep{WangLNLK2024} (\eg, CHAMMI-CP) with both real and synthetic asymmetric and symmetric noise (VLCS and PACs for real noise, and DomainNet, Officehome, and TerraIncognita for synthetic).  These datasets provide results over a ride range of applications, \eg, VLCS and DomainNet are typical web images, CHAMMI-CP contains cell images, and TerraIncognita contains images from wildlife cameras.

\noindent\textbf{Metrics.} All of our datasets report top-1 classification accuracy on both ID and OOD data.  In some experiments we also report the average of these two metrics to measure overall \taskAbbr{} performance. 

\textbf{Experimental setup.} As shown in \Cref{fig:related_work}, there are three components to evaluating \taskAbbr{}: in-domain (ID) performance, out-of-domain (OOD) generalization, and robustness to noise.  Our experiments are designed to evaluate each component by reporting both ID and OOD accuracy using a "leave-one-out" protocol used by domain generalization benchmarks like DomainBed~\citep{gulrajani2020search-erm}.  In other words, we train on all domains but one, compute accuracy on the test sets for both ID and OOD data, and average the results. Robutness to label noise is measured using real-world noise (\ie, noise that appears naturally) as well as controlled experiments like those we conducted using RotatedMNIST in \Cref{sec:nag} and \Cref{sec:DL4ND}.  Real-world noise experiments validate that our approach works in practice, while the controlled experiments enable us to provide into deeper insights by controlling the noise that appears in the datasets. 

Following~\citet{ballas2025gradient,gulrajani2020search-erm,Wang_2025_CVPR}, our experiments use a ResNet-50~\citep{he2016deep} pretrained on ImageNet~\citep{deng2009imagenet}, except for CHAMMI-CP which, following \citet{chen2024chammi}, uses a ConvNeXt~\citep{liu2022convnet} pretrained on ImageNet 22K.  
While some prior work has used large multimodal models (LMMs) in their experiments,  LMMs have applications restrictions due to task or computational requirements. \Eg,  CHAMMI-CP~\citep{chen2024chammi} has up to 5 channel cell images, so LMMs that process 3 channel natural images do not apply. Additionally, recent work has shown the high performance of LMMs on DG benchmarks like VLCS and OfficeHome is largely due to train/test contamination~\citep{teterwak2024large}, which is significantly less pronounced in TerraIncognita.  This also likely explains why some of our baseline methods outperform LLMs like GPT-4v (by up to 15\%) reported in \citet{han2024how} on TerraIncognita using a ResNet-50. Additional implementation details are in \Cref{app:implementation_details}.

\subsection{Results}
\label{sec:real-world-results}

\begin{figure*}[t]
  \centering
  \begin{subfigure}[b]{0.49\linewidth}
    \includegraphics[width=\textwidth,trim=0cm 3.7cm 11.1cm 1.2cm,clip]{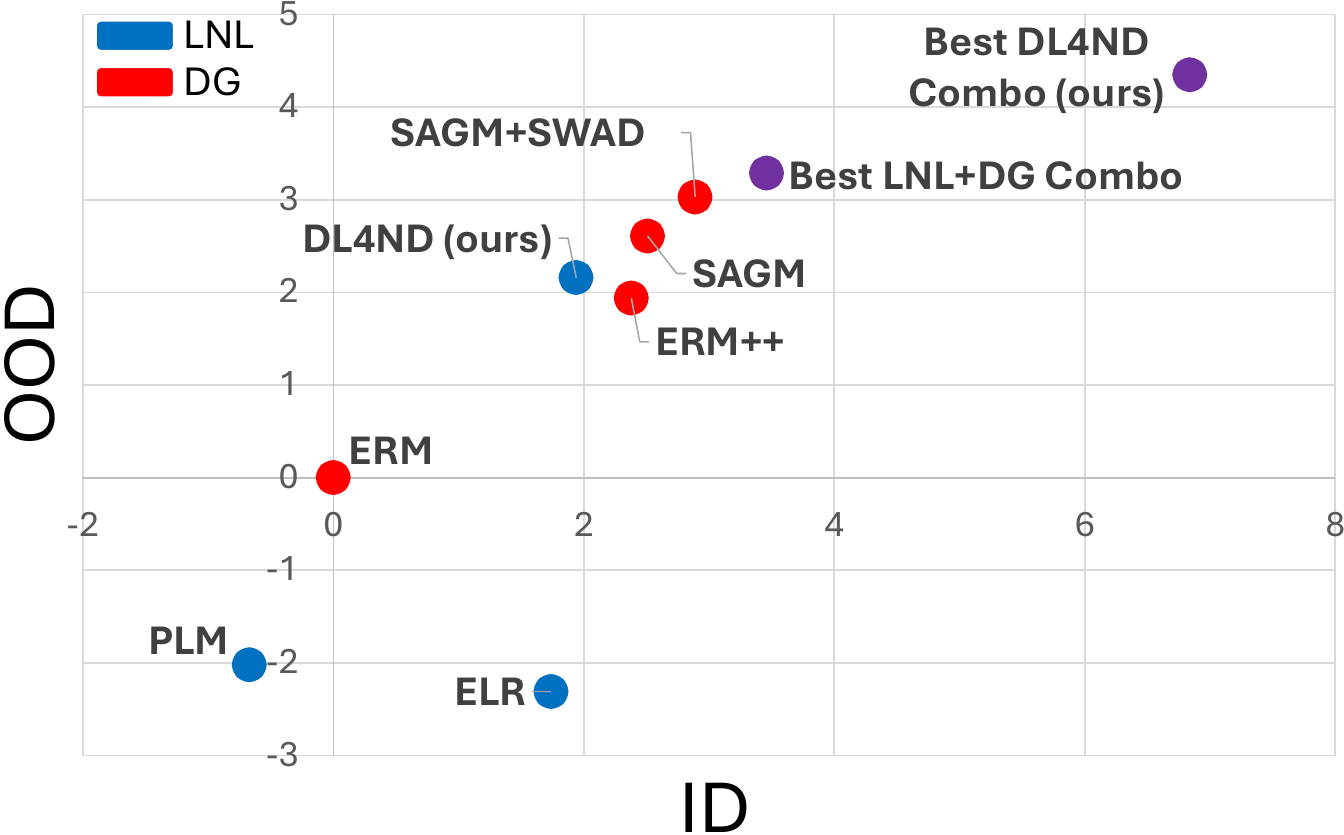}
    \caption{VLCS~\citep{fang2013unbiased}}
    \end{subfigure}
    \hfill
    \begin{subfigure}[b]{0.49\linewidth}
    \includegraphics[width=\textwidth,trim=0cm 3.7cm 11.1cm 1.2cm,clip]{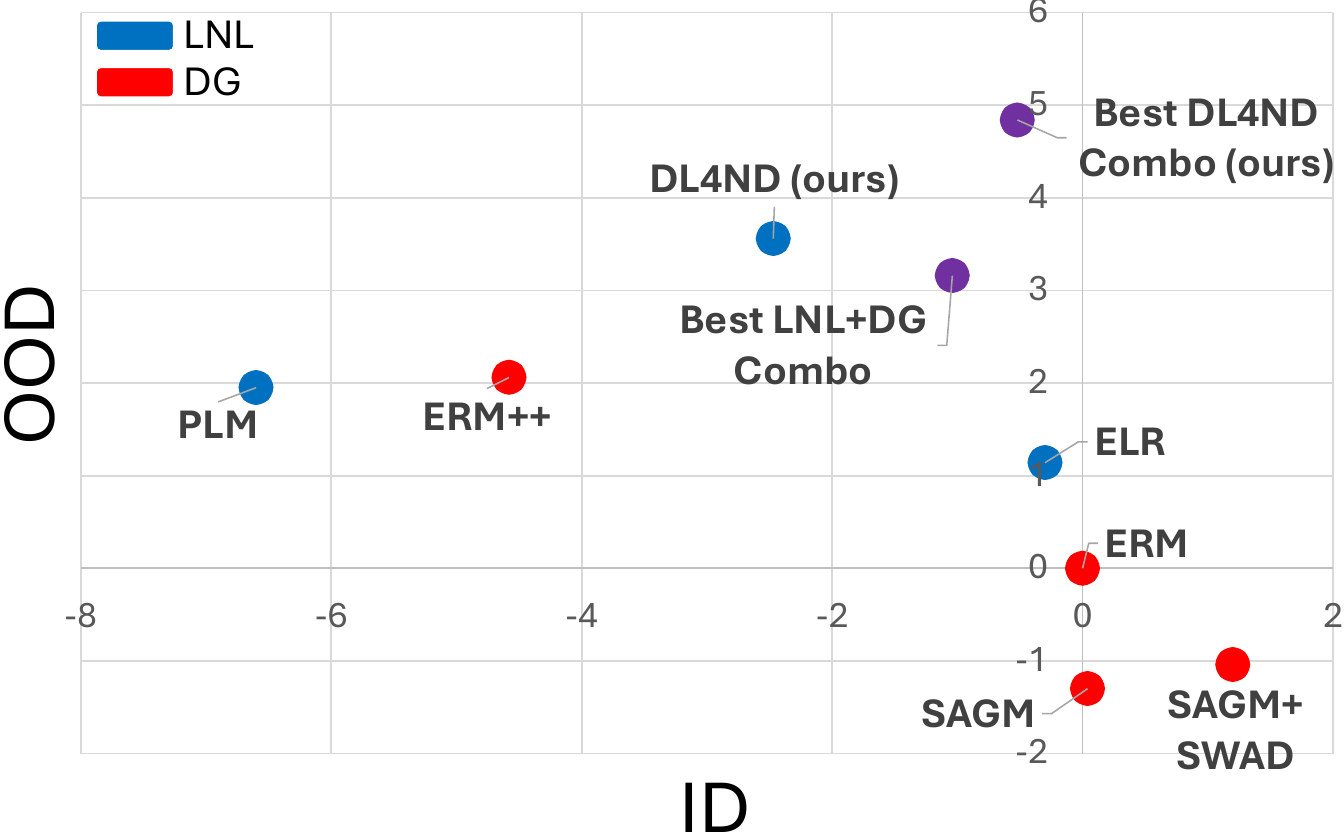}
    \caption{CHAMMI-CP~\citep{chen2024chammi}}
    \end{subfigure}
    \vspace{-2mm}
    \caption{Summary of real-world noise top-1 accuracy changes on VLCS~\citep{fang2013unbiased} and CHAMMI-CP~\citep{chen2024chammi}. We center results around ERM~\citep{gulrajani2020search-erm}, and report relative absolute changes in both in-domain (ID) and out-of-domain (OOD) performance for the best LNL (ELR~\citep{liu2020early}, PLM~\citep{zhao2024estimating}) and DG methods (ERM++~\citep{teterwak2023erm++}, SAGM~\citep{wang2023sharpness}, SWAD~\citep{cha2021swad}). \Cref{app_sec:extended_results} contains additional methods and combinations. We find \methodAbbrspace outperforms prior LNL methods, even when combined with DG approaches. See \Cref{sec:real-world-results} for further discussion.}
  \label{fig:real-world}
  %\vspace{-2mm}
\end{figure*}

\begin{table}[t]
    \begin{minipage}{.38\linewidth}
  \centering
  \setlength{\tabcolsep}{2pt}
  %\small
    \caption{Real-world noise top-1 accuracy on PACS~\citep{Li2017DeeperBA}. Integrating \methodAbbrspace with DG methods provides up to a 2\% average gain.}
    %\vspace{-2mm}
      \label{table:pacs}
   \begin{tabular}{lccc}
    \toprule
    Method   & ID & OOD &AVG \\
    \midrule
    SAGM &96.3& 85.3 & 90.8 \\
    \hspace{2mm} w/ DL4ND (ours) & \textbf{\underline{97.3}} & \underline{88.8} & \underline{93.1} \\
    ERM++ & \underline{96.7} & 89.2 & 92.9 \\
    \hspace{2mm} w/ DL4ND (ours) & 96.5 & \textbf{\underline{90.1}} & \textbf{\underline{93.3}} \\
    \bottomrule
  \end{tabular}

    \end{minipage}
    \hfill
    \begin{minipage}{.58\linewidth}
      \centering
  \setlength{\tabcolsep}{2pt}
  %\small
    \caption{Results with 60\% symmetric noise on OfficeHome~\citep{venkateswara2017deep} (see \Cref{table:synthetic_noise} for asymmetric noise).  Using \methodAbbrspace provides up to 12.5\% gain.}
    %\vspace{-2mm}
      \label{table:symm}
   \begin{tabular}{lccc}
    \toprule
    Method   & ID & OOD &AVG \\
    \midrule
    ERM & 45.8 & 40.5 & 43.2 \\
    \hspace{2mm} w/ DL4ND (ours) & \underline{47.9} & \underline{49.9} & \underline{48.9} \\
    SAGM & 48.6 & 40.3 & 44.4 \\
    \hspace{2mm} w/ DL4ND (ours) &  \underline{52.0} & \underline{52.6} & \underline{52.2} \\
    ERM++ &  56.7 & 48.7 & 52.7 \\
    \hspace{2mm} w/ DL4ND (ours) & \textbf{\underline{60.3}} & \textbf{\underline{59.4}} & \textbf{\underline{59.8}} \\
    \bottomrule
  \end{tabular}
    \end{minipage} 
\end{table}

\Cref{fig:real-world} summarizes key results comparing our \methodAbbr{} approach with top methods from the LNL and DG literature, along with their best combinations, on  VLCS~\citep{fang2013unbiased} and CHAMMI-CP~\citep{chen2024chammi} (additional results in \Cref{app_sec:extended_results}).  We make two notable observations.  First, our \methodAbbr{} method on its own outperforms other LNL methods, and  \methodAbbr{} is the only approach on CHAMMI-CP that provides an average boost on its own.  Second, \methodAbbr{} combined with DG methods outperforms combinations of prior work in DG+LNL by 1-2\% on average. \Cref{table:pacs} demonstrates that the advantage from \methodAbbr{} also extends to PACS~\citep{Li2017DeeperBA}, booting performance when combined with DG methods by up to 2\%.  These results show that \methodAbbr{} is beneficial across diverse datasets with real-world noise, especially when combined with DG methods.

\Cref{table:symm} and \Cref{table:synthetic_noise} provides results on a controlled experiments where we add symmetric noise to OfficeHome~\citep{venkateswara2017deep} and asymmetric noise to OfficeHome, TerraIncognita~\citep{beery2018recognition}, and DomainNet~\citep{peng2019moment}, respectively.  Overall, \methodAbbr{} results in a boost to performance in most settings, obtaining the best performance in 11 of 13 cases, with gains as large as 12.5\% (OOD results on symmetric OfficeHome). We find that as we increase the amount of noise in a dataset that the benefit of \methodAbbr{} also grows.

\Cref{table:ablations} provides an ablation study that shows each component of \methodAbbr{} improves performance.  Specifically, \emph{w/o relabel} refers to removing rather than relabeling samples, \emph{w/o cross-domain} uses within-domain rather than cross-domain comparisons to identify noise (described in \Cref{sec:cross_domain}), and \emph{w/o small-loss proxy} reports results that uses all samples to create the proxies for cross-domain comparisons rather than just low-loss samples (discussed in ~\Cref{sec:low_loss}). As shown in \Cref{table:ablations}, each component provides a 2-4\% gain in most settings, with the best performing method on each dataset using all components.  In \Cref{table:relabeling} we also show that the gain from cross-domain comparisons can at least be partly explained as an improvement in the precision of relabeled samples.

\begin{table*}[t]
    \centering
    \setlength{\tabcolsep}{1.2pt} % Reduce column separation
    %\scriptsize % Use smaller font
    %\small
    \caption{Asymmetric noise results on OfficeHome~\citep{venkateswara2017deep}, TerraIncognita~\citep{beery2018recognition}, and DomainNet~\citep{peng2019moment}.  We \textbf{bold} overall best results, and \underline{underline} best results between \methodAbbrspace and a specific DG method it is integrated with.  \methodAbbrspace boosts performance in 8 of 10 settings, with a 9\% gain on DomainNet.  More results are in \Cref{app_table:synethic_erm}.}
    %\vspace{-2mm}
    \begin{tabular}{l|cccccc|cccccc|cc}
        \toprule
        & \multicolumn{6}{c|}{\textbf{OfficeHome}} & \multicolumn{6}{c}{\textbf{TerraIncognita}} & \multicolumn{2}{|c}{\textbf{DomainNet}}\\
        \midrule
        Method & \multicolumn{2}{c}{No Noise} & \multicolumn{2}{c}{20\% Noise} & \multicolumn{2}{c|}{40\% Noise} & \multicolumn{2}{c}{No Noise} & \multicolumn{2}{c}{20\% Noise} & \multicolumn{2}{c|}{40\% Noise} & \multicolumn{2}{c}{40\% Noise}\\
        \cmidrule(lr){2-3} \cmidrule(lr){4-5} \cmidrule(lr){6-7} \cmidrule(lr){8-9} \cmidrule(lr){10-11} \cmidrule(lr){12-13} \cmidrule(lr){14-15}
        & ID & OOD & ID & OOD & ID & OOD & ID & OOD & ID & OOD & ID & OOD & ID & OOD\\
        \midrule
        SAGM & 83.4 & 69.1 & 76.7 & 64.0 & 62.0 & 52.0 & 86.7 & 51.3 & \textbf{\underline{78.2}} & 40.4 & 56.4 & 30.9 & 43.6 & 26.9\\
        \hspace{2mm} w/ DL4ND (ours) & -- & -- & \textbf{\underline{81.4}} 
                        & \underline{66.6} 
                        & \textbf{\underline{69.3}} 
                        & \underline{55.0} 
                & -- & --& 77.1 
                        & \underline{43.5} 
                        & \textbf{\underline{58.7}} 
                        & \underline{32.8} & \textbf{\underline{52.8}} & \underline{35.6}\\
        ERM++ & 85.1 & 71.7 & \underline{78.5} & 65.8 & \underline{63.1} & 52.3 & 85.9 & 47.5 & \underline{75.9} & \textbf{\underline{45.0}} & 56.1 & 30.6 & 42.4 & 28.9\\
        \hspace{2mm} w/ DL4ND (ours) & -- & -- & 76.6 
                               & \textbf{\underline{68.6}} 
                               & 62.7 
                               & \textbf{\underline{56.4}} 
                      & -- & --& 73.0 
                               & 43.2 
                               & \underline{57.0} 
                               & \textbf{\underline{37.5}} & \underline{51.1} & \textbf{\underline{36.2}}\\
        \bottomrule
    \end{tabular}
    \label{table:synthetic_noise}
    %\vspace{-2mm}
\end{table*}

\begin{table}[t]       
  \centering
  \setlength{\tabcolsep}{3pt}
  \caption{Ablation study on VLCS~\citep{fang2013unbiased} and CHAMMI-CP~\citep{chen2024chammi}.  We \textbf{bold} overall best results, and \underline{underline} best results between when \methodAbbrspace is integrated and its ablations.  We show our model components boost performance by 2-4\%.}
    %\vspace{-2mm}
   \begin{tabular}{lcccccc}
    \toprule
    Method  & \multicolumn{3}{c}{VLCS} & \multicolumn{3}{c}{CHAMMI-CP} \\
    \cmidrule(r){2-4} \cmidrule(r){5-7} 
     & ID & OOD &AVG & ID & OOD &AVG\\
    \midrule
   SAGM+SWAD w/DL4ND (ours)  & \underline{91.9} & \underline{88.6} & \underline{90.6} & \textbf{\underline{76.6}} & 47.3 & \textbf{\underline{61.9}}\\
    \hspace{3mm} w/o relabel &88.5 & 87.2& 87.8 & 70.9 & 47.2 & 59.0\\
    \hspace{3mm} w/o cross-domain & 88.6 & 87.8 & 88.2& 68.5 & \textbf{\underline{49.0}} & 58.8\\
    \hspace{3mm} w/o small-loss proxy & 88.2 & 87.0& 87.6 & 73.6 & 45.8& 59.7\\
    \midrule
    ERM++ w/DL4ND (ours)  & \textbf{\underline{95.3}} & \textbf{\underline{89.0}} & \textbf{\underline{92.2}} & \underline{72.9} & 
    44.3 & 58.6 \\
    \hspace{3mm} w/o relabel &89.5 & 88.8& 89.2& 72.5 & 45.4& \underline{58.9} \\
   \hspace{3mm} w/o cross-domain  & 89.5 & 87.6 & 88.6 & 69.7 & \underline{47.3} & 58.5\\
    \hspace{3mm} w/o small-loss proxy  & 93.5& 87.7&  90.6&  68.9 & 43.4 & 56.1\\
    \bottomrule
  \end{tabular}
  %\vspace{-2mm}
  \label{table:ablations}
\end{table}

\begin{table}[t]  
\centering
\begin{minipage}[c]{0.5\textwidth}
\caption{Comparing relabeling precision at two noise levels on OfficeHome~\citep{venkateswara2017deep} and TerraIncognita~\citep{beery2018recognition}.  Our cross-domain comparisons improves precision by up to 10\% over using a single domain.}
    \label{table:relabeling}
\end{minipage}
\hspace{0.2cm}
\begin{minipage}[c]{0.4\textwidth}
\setlength{\tabcolsep}{3pt}
\begin{tabular}{lcccc}
    \toprule
    Comparison  & \multicolumn{2}{c}{OfficeHome} & \multicolumn{2}{c}{TerraIncognita} \\
    \cmidrule(r){2-3} \cmidrule(r){4-5} 
    & 20\% & 40\% & 20\% & 40\%\\
    \midrule
    same-domain & 99.5 & 94.0 & 82.5  & 64.3  \\
    cross-domain & \textbf{99.8} & \textbf{97.2} & \textbf{86.7}  & \textbf{74.3}  \\
    \bottomrule
  \end{tabular}
%\vspace{10pt}
%\vspace{-2mm}
\end{minipage}%
%\vspace{-2mm}
\end{table}

\subsection{Why don't naive LNL+DG combinations perform better?}
\label{sec:LNL+DG}

\begin{figure}[t]
\centering
\begin{minipage}[c]{0.55\textwidth}
\includegraphics[width=\linewidth]{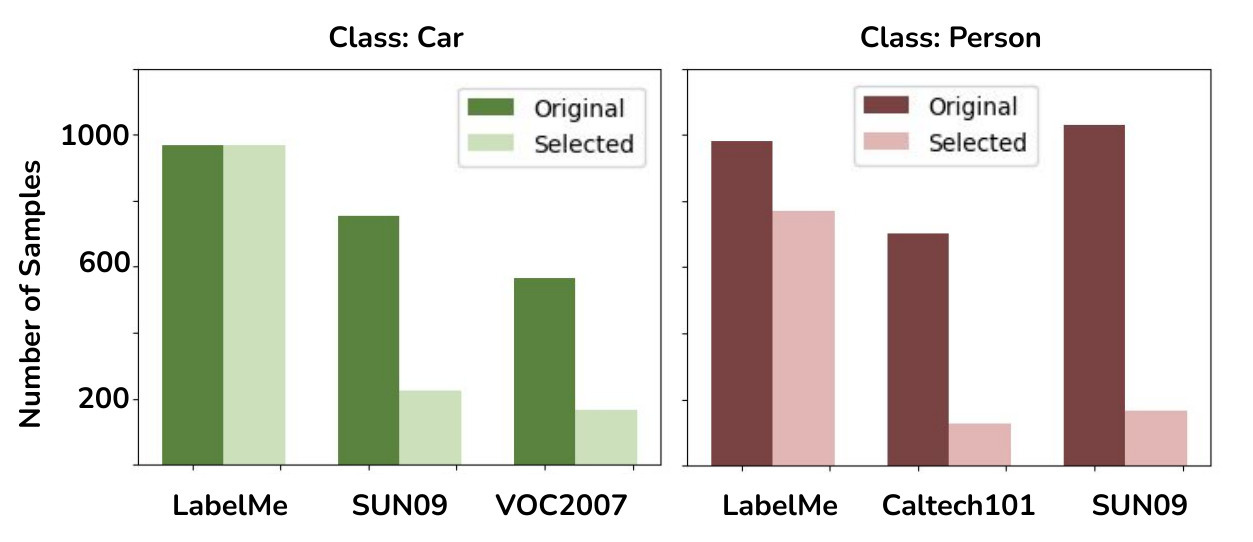}
%\vspace{10pt}
%\vspace{-2mm}
\end{minipage}%
%\hspace{0.3cm}
\hfill
\begin{minipage}[c]{0.43\textwidth}
\caption{\textbf{Changes in domain distribution after the UNICON sample selection process on VLCS}~\citep{fang2013unbiased}. (\textit{Left bar: number of samples before selection, right bar: after selection}). These two cases illustrate a risk of skewing domain distributions from the LNL selection process. See Sec.~\ref{sec:LNL+DG} for more details.}
    \label{fig:sample_selection}
\end{minipage}
%\vspace{-2mm}
\end{figure}

 \begin{table*}[h!]       
  \centering
  \setlength{\tabcolsep}{3pt}
    \caption{Comparison of relabeling methods on real-world noise top-1 accuracy on VLCS~\citep{fang2013unbiased} and CHAMMI-CP~\citep{chen2024chammi}.  We use the best performing DG method SAGM~\citep{wang2023sharpness}+SWAD~\citep{cha2021swad} and combine them with to relabeling method UNICON~\citep{karim2022unicon} or \methodAbbr.  \emph{Per-domain sampling} uses domain labels to separate samples by their sources, and relabels samples evenly across domains.  This effectively reduces to the standard LNL task addressed by UNICON, since each source does not contain domain shifts. However, \methodAbbrspace still performs better, especially on VLCS. See \Cref{sec:LNL+DG} for discussion.}
    % \vspace{-2mm}
   \begin{tabular}{lcccccccc}
    \toprule
    DG Method & LNL Method & Per-domain   & \multicolumn{3}{c}{VLCS} & \multicolumn{3}{c}{CHAMMI-CP} \\
    \cmidrule(r){4-6} \cmidrule(r){7-9} 
       & & sampling& ID & OOD &AVG & ID & OOD &AVG\\
    \midrule
    --& UNICON & -- & 89.9 & 84.0 & 86.9 & 76.7 & 42.0 & 59.4\\
     SAGM+SWAD  & -- & -- & 91.4 & 87.7 & 89.5 & \textbf{78.3} & 41.5 & 59.9 \\
    SAGM+SWAD & UNICON & --  & 87.8 & 82.8 & 85.3 & 75.4 & 43.3 & 59.3\\
     SAGM+SWAD & UNICON & \checkmark & 89.3 & 86.8 & 88.1 & 77.0 & 46.3 & 61.6\\
    SAGM+SWAD & DL4ND (ours) & --  & \textbf{91.9} & \textbf{88.6} & \textbf{90.3} & 76.6 & \textbf{47.3} & \textbf{61.9}\\
    \bottomrule
  \end{tabular}
 % \vspace{-2mm}
  \label{table:naive_combo}
\end{table*}

At a high level, \methodAbbr{} resembles UNICON~\citep{karim2022unicon}, which also estimates and relabels potential noisy samples.  However, as discussed in the Introduction, LNL methods like UNICON assume that high-loss samples tend to be noise, but in \taskAbbr{} these samples may also be domains that are harder to learn. For example, \Cref{fig:sample_selection} shows that samples selected as part of the clean subset when training on VLCS mostly come from LabelMe, whereas some domains like SUN09 are mostly identified as noise.  This makes it challenging to learn from all domains as some may have many samples incorrectly relabeled (especially those along the decision boundary).  In effect, this relabeling introduces more noise rather than fixing it.  \Cref{fig:vlcs_clean_size} in the Appendix shows these issues persist even when decreasing ratio of samples allowed to be relabeled.  This observation is similar to the one that UNICON made of DivideMix~\citep{li2020dividemix}, where they noted that some categories were more prone to relabeling than others and argued for more balanced approach to address it.

In \Cref{table:naive_combo} we extend this idea to the multi-domain setting, where we ensure relabeling is also source-balanced.  Specifically, we perform per-domain sampling, where we relabel only a subset of each domain (\ie, we relabel 20\% of the samples in each domain rather than 20\% of the dataset regardless of domain).  \Cref{table:naive_combo} shows this does improve the performance of combining UNICON with DG methods SAGM+SWAD by 2-3\% on average.   However, it only improves performance compared to SAGM+SWAD alone on CHAMMI-CP, and actually decreases performance on VLCS. This is likely due, in part, to the fact that in this setting UNICON will still utilize same-domain comparisons, which we showed in the previous section underperforms \methodAbbr's cross-domain approach.   Thus, as seen comparing the last two lines of \Cref{table:naive_combo}, our approach is 2\% better on VLCS while also providing a small gain on CHAMMI-CP.  Notably, \methodAbbr{} improves performance over SAGM+SWAD alone on both datasets unlike UNICON.

\section{Conclusion}
% In this work, we address the complexities of real-world data, which is often noisy and diverse, posing significant challenges for model generalization. Our task, \task, focuses on training models capable of handling both in-domain noise and out-of-domain generalization. To this end, we proposed a unified framework that approaches the problem from both Learning with Noisy Labels (LNL) and Domain Generalization (DG) perspectives. We examined the noise ratios in three real-world datasets, providing a comprehensive set of experiments related to in-domain noise and domain shifts, enabling us to evaluate \taskspace across a broad spectrum.
%the interactions between LNL and DG methods across a broad spectrum.

This work addresses the challenges of training noisy, diverse real-world data by exploring \taskspace (\taskAbbr), which focuses on handling in-domain noise and improving out-of-domain generalization.
We highlight several key takeaways. First, \taskAbbrspace presents new challenges, which complicate the task of distinguishing between noise and domain distribution shifts. Second, a naive combination of LNL and DG does not effectively address this task. Domain shift can interfere with noise detection, and LNL-based sample selection can inadvertently skew the domain distribution. Lastly, we demonstrate that using cross-domain comparisons as a critical signal for noise detection significantly improves performance. Noise, which lacks the intrinsic class features, fails to exhibit closer distances to other domains. Experimental results validate the effectiveness of our approach, and the discussion also provides insights for further advancing \taskAbbr.

\textbf{Acknowledgments.}
This study was supported by the National Science Foundation under NSF-DBI award 2134696. Any opinions, findings, and conclusions or recommendations expressed in this material are those of the author(s) and do not necessarily reflect the views of the supporting agency.

\section*{Ethics Statement}

 This paper addresses \taskAbbr, a task that requires models to perform well on both in-domain and out-of-domain data when training on datasets with label noise.  This can result in models that can effectively learn from a wide variety of data, including cell painting data where prior work in tasks like LNL found especially challenging due to its high amounts of label noise~\citep{WangLNLK2024}.  However, like many topics in this field, also can enable bad actors to use these models to train more effective recognition systems for nefarious purposes.  Additionally, users should be mindful that although we provide an evaluation on a diverse set of datasets, they still make mistakes in their predictions that may vary depending on the dataset.  Thus, researchers and engineers should be mindful of these factors when deploying a system for end-users.

\section*{Reproducibility Statement}

We have released our code and data to ensure it can be reproduced.  We provide detailed implementation details as well as an in-depth example of how to integrate the methods in our experiments in \Cref{app:implementation_details}.  Our released code is capable of training and testing the models we compared to in a unified codebase.  This enables additional methods to be easily integrated and the data loaders required to evaluate models on our benchmarks.  Further, we have also included some pretrained models for ease of use.

\bibliography{main}
\bibliographystyle{iclr2026_conference}

\appendix
%\section{Appendix}
%\appendix
% \begin{center}
%     \Large \bfseries \task: Robustness to In-Domain Noise and Out-of-Domain Generalization Supplementary
% \end{center}

\section{Extended Results}
\label{app_sec:extended_results}

\Cref{table:real-world} contains detailed results for both DG and LNL-only methods.  Generally speaking, the results follow intuition, where DG methods perform better on VLCS, which has less noise and more dramatic domain shifts.  In contrast, \methodAbbrspace being an LNL method performs the best on CHAMMI-CP, which has more noise and smaller shifts.  That said, our approach does outperform other LNL methods on VLCS by 2\%.  Notably, for DG methods the more regularization focused methods like SWAD are necessary to get best performance.  This observation is key as in the LNL methods ELR is also a regularization-style approach and performs similarly to \methodAbbrspace on CHAMMI-CP.  Thus, conceptually our approach has more differences than ELR compared to the DG methods that do well on our \taskAbbrspace task.

\begin{table*}[t]       
  \centering
  \caption{Comparing methods of addressing \taskAbbrspace on VLCS~\citep{fang2013unbiased} and CHAMMI-CP~\citep{chen2024chammi} either via (a) by using a method from prior work in the DG literature or (b) LNL methods like our approach \methodAbbrspace approach.  We bold the best performance overall and underline the best LNL method.  We find \methodAbbrspace performs best among LNL methods, and does especially well on CHAMMI-CP. See \Cref{app_sec:extended_results} for discussion.}
   \begin{tabular}{llcccccc}
    \toprule
    & Method  & \multicolumn{3}{c}{VLCS} & \multicolumn{3}{c}{CHAMMI-CP} \\
    \cmidrule(r){3-5} \cmidrule(r){6-8} 
    &   & ID & OOD &AVG & ID & OOD &AVG\\
    \midrule
    (a) & ERM~\citep{gulrajani2020search-erm}  & 88.5 & 84.6 & 86.6 & 77.1  & 42.5 & 59.8\\
     & VREx~\citep{krueger2021out}  & 89.0 & 84.4 & 86.7  & 74.8 & 44.8 & 59.8 \\
    & SWAD~\citep{cha2021swad}  & 90.8 & 86.2 & 88.5 & 73.9 & 43.7 & 58.8\\
    & Fishr~\citep{rame2022fishr}  & 88.9 & 85.8 & 87.4 & 73.9  & 44.0 & 59.0\\
    & MIRO~\citep{cha2022miro}  & 88.9 & 83.8 & 86.4 & 65.5 & \textbf{46.6} & 56.0\\
    & SAGM~\citep{wang2023sharpness} & 91.0 & 87.2 & 89.1 & 77.1 & 41.2 & 59.2 \\
    & DISAM~\cite{zhang2024domain}  & 89.6 & 84.9 & 87.2 &72.4 & 44.8 & 58.6 \\
    & ERM++~\citep{teterwak2023erm++}  & 90.9 & 86.6 & 88.7 & 72.5 & 44.6 & 58.5 \\
    % & URM~\citep{krishnamachari2024uniformly}  & \textcolor{red}{84.28} &\textcolor{red}{76.77} & \textcolor{red}{80.53}&  &  &  \\
    % & GGA~\citep{ballas2025gradient}  & \textcolor{red}{84.63} &\textcolor{red}{77.39} & \textcolor{red}{81.01}&  &  &  \\
    % & GGA (original noisy VLCS)  & \textcolor{red}{85.57} &\textcolor{red}{76.94} & \textcolor{red}{81.26}&  &  &  \\
    % & GGA-L  & \textcolor{red}{84.06} &\textcolor{red}{76.60} & &  &  &  \\
    % & GGA-L (original noisy VLCS)  & \textcolor{red}{84.08} &\textcolor{red}{75.47} & &  &  &  \\
    & MIRO+SWAD & 88.9 & 83.7 & 86.3 & 67.3 & 45.8 & 56.6\\
    & SAGM+SWAD  & \textbf{91.4} & \textbf{87.7} & \textbf{89.5}& \textbf{78.3} & 41.5 & 59.9 \\
    \midrule
	(b) & ELR~\citep{liu2020early}  &90.3 & 82.3 & 86.3  & \underline{76.8} & 43.6 & 60.2\\
    & UNICON~\citep{karim2022unicon}  & 89.9 & 84.0& 86.9 & 76.7 & 42.0 & 59.4\\
	& DISC~\citep{li2023disc}  & 88.7 & 82.5& 85.6 & 43.3 & 41.3 & 42.3\\
    & PLM~\citep{zhao2024estimating}  & 87.9 & 82.6 & 85.2 & 70.5& 44.4 & 57.5\\
    & \methodAbbrspace (ours)  & \underline{90.5} & \underline{86.8} & \underline{88.6} & 74.6 & \underline{46.0} & \underline{\textbf{60.3}} \\
    \bottomrule
  \end{tabular}
  \label{table:real-world}
\end{table*}

As shown in \Cref{table:real-world_combos}, this conceptual difference manifests itself when we aim to combine DG and LNL methods.  Notably, while ELR actually performs worse on CHAMMI-CP, suggesting that the model becomes too constrained with the additionl of multiple types of regularization, \methodAbbrspace performs better, getting a 2\% boost on CHAMMI and a 3.5\% gain on VLCS when combined with DG method (comparing the last line of \Cref{table:real-world} to the last two lines of \Cref{table:real-world_combos}).  The effectiveness of the DG methods also changes when combined with \methodAbbr, as SAGM+SWAD performed best in \Cref{table:real-world} on VLCS, but we get nearly a 3\% gain combining ERM++ with \methodAbbrspace in \Cref{table:real-world_combos}.

\begin{table*}[t]       
  \centering
  \caption{Comparing different combinations of DG methods on VLCS~\citep{fang2013unbiased} and CHAMMI-CP~\citep{chen2024chammi} with (a) LNL methods from prior work or (b) our \methodAbbrspace approach.  We find combinations with our \methodAbbrspace method gets best performance. See \Cref{app_sec:extended_results} for discussion.}
   \begin{tabular}{lcccccccc}
    \toprule
    & DG Method & LNL Method  & \multicolumn{3}{c}{VLCS} & \multicolumn{3}{c}{CHAMMI-CP} \\
    \cmidrule(r){4-6} \cmidrule(r){7-9} 
    & &  & ID & OOD &AVG & ID & OOD &AVG\\
    \midrule
    (a) & ERM++ & ELR  & 89.7 & 85.4 & 87.6 & 75.7 & 42.0 & 58.9\\
     &MIRO+SWAD & ELR  &91.5 & 86.7 & 89.1  &70.7 & 44.8 & 57.8\\
     &MIRO & ELR  & 90.8 & 84.5 & 87.7 & 74.5 & 41.3 & 57.9\\
	 &SWAD & ELR  & 92.0& 87.9 & 90.0 & 73.5& 44.7 & 59.1\\
     &MIRO & UNICON  & 89.8 & 83.4& 86.6 & \textbf{77.0} & 43.4 & 60.2 \\
	 &MIRO+SWAD & UNICON  & 88.9 & 83.7 & 86.3 & 76.0 & 45.7 & 60.8\\
     & SAGM+SWAD & UNICON  & 87.8 & 82.8 & 85.3 & 75.4 & 43.3 & 59.3\\
    \midrule
    (b) & VREx  & DL4ND (ours) & 91.2 & 87.0 & 89.1 &75.3 & 46.7 & 61.0 \\
    & Fishr & DL4ND (ours)  & 89.9 & 86.5 & 88.2 &73.8 & 46.1 & 60.0 \\
    & MIRO & DL4ND (ours)  & 93.5 & 86.7 & 90.1 &70.4 &46.7 & 58.5  \\
    & MIRO+SWAD & DL4ND (ours)  &  91.7 & 88.0 & 89.9  &71.2 & 46.6 & 58.9  \\
    & SAGM & DL4ND (ours)  & 91.9 & 88.4 & 90.1 &76.2 & 46.6 & 61.4 \\
    & SAGM+SWAD & DL4ND (ours)  & 91.9 & 88.6 & 90.3 &76.6 & \textbf{47.3} & \textbf{61.9}\\
    & ERM++ & DL4ND (ours)  & \textbf{95.4} & \textbf{89.0} & \textbf{92.2} &72.9 & 44.3 & 58.6 \\
    %& URM~\citep{krishnamachari2024uniformly} + DL4ND (ours)  & \textcolor{red}{85.54} &\textcolor{red}{77.74} & \textcolor{red}{81.64}&  &  &  \\
    \bottomrule
  \end{tabular}
  \label{table:real-world_combos}
\end{table*}

In addition, as we discussed in \Cref{sec:LNL+DG}, part of the reason why UNICON does not perform as well as \methodAbbrspace in \Cref{table:real-world_combos} is due to the fact that it tends to be biased towards detecting samples from a subset of domains as noise.  \Ie, harder to learn domains incorrectly get identified as largely noise.  \Cref{table:extended_domainlabel} expands on our results from the main paper where we assume we are provided with domain labels, which enables us to per-domain sampling.  On VLCS all the UNICON results underperforms \methodAbbrspace by 2\%. The various combinations of UNICON with DG methods generally perform the same (even though they do all outperform combinations in \Cref{table:real-world_combos}), with the only real variation combing on CHAMMI-CP.  However, even on this dataset our \methodAbbrspace reports a slight advantage, mostly due to stronger OOD performance.

\begin{table*}[t]       
  \centering
    \caption{Comparison of relabeling methods on real-world noise top-1 accuracy on VLCS~\citep{fang2013unbiased} and CHAMMI-CP~\citep{chen2024chammi} expanding on our results from \Cref{sec:LNL+DG}. \emph{Per-domain sampling} uses domain labels to separate samples by their sources, and relabels samples evenly across domains. \methodAbbrspace  outperforms UNICON in nearly all cases.}
   \begin{tabular}{lcccccccc}
    \toprule
    DG Method & LNL Method & Per-domain   & \multicolumn{3}{c}{VLCS} & \multicolumn{3}{c}{CHAMMI-CP} \\
    \cmidrule(r){4-6} \cmidrule(r){7-9} 
       & & sampling& ID & OOD &AVG & ID & OOD &AVG\\
    \midrule
    MIRO & UNICON & \checkmark &91.2 & 85.8& 88.5 & 76.9 & 45.2 & 61.1\\
	MIRO+SWAD & UNICON & \checkmark & 90.6& 86.0 & 88.3 & 76.5 & 43.6 & 60.0\\
    SAGM+SWAD & UNICON & \checkmark & 89.3 & 86.8 & 88.1 & \textbf{77.0} & 46.3 & 61.6\\
    SAGM+SWAD &  DL4ND (ours) & -- & \textbf{91.9} & \textbf{88.6} & \textbf{90.3} & 76.6 & \textbf{47.3} & \textbf{61.9}\\
    \bottomrule
  \end{tabular}
  \label{table:extended_domainlabel}
\end{table*}

\Cref{app_table:synethic_erm} reports an ERM baseline~\citep{gulrajani2020search-erm} on its own and when combined with our approach on OfficeHome~\citep{venkateswara2017deep} and TerraIncognita~\citep{beery2018recognition} to  supplement the results in the main paper.  We find that \methodAbbrspace improves performance by 3-18\%, demonstrating that our approach provide a significant benefit over DG methods alone.

\begin{table*}[t]
    \centering
    %\scriptsize % Reduce font size
    \setlength{\tabcolsep}{3pt} % Reduce column separation
      \caption{Synthetic noise results on OfficeHome~\citep{venkateswara2017deep} and TerraIncognita~\citep{beery2018recognition} to supplement results from \Cref{sec:real-world-results}.  \methodAbbrspace boosts performance by 3-18\% over ERM~\citep{gulrajani2020search-erm} alone.}
    \begin{tabular}{l|cccccc|cccc}
        \toprule
        & \multicolumn{6}{c|}{\textbf{OfficeHome}} 
        & \multicolumn{4}{c}{\textbf{TerraIncognita}} \\ 
        %& \multicolumn{4}{c}{\textbf{DomainNet}} \\
        \cmidrule(lr){2-7} \cmidrule(lr){8-11} %\cmidrule(lr){12-15}
        Method & \multicolumn{2}{c}{No Noise} & \multicolumn{2}{c}{20\% Noise} & \multicolumn{2}{c|}{40\% Noise} 
        & \multicolumn{2}{c}{No Noise} & \multicolumn{2}{c}{40\% Noise}\\
        %& \multicolumn{2}{c}{No Noise} & \multicolumn{2}{c}{0.4 Noise} \\
        \cmidrule(lr){2-3} \cmidrule(lr){4-5} \cmidrule(lr){6-7}
        \cmidrule(lr){8-9} \cmidrule(lr){10-11} %\cmidrule(lr){12-13} \cmidrule(lr){14-15}
        & ID & OOD & ID & OOD & ID & OOD & ID & OOD & ID & OOD \\%& ID & OOD & ID & OOD \\
        \midrule
        ERM%~\citep{vapnik1999overview}
            & 80.6 & 65.6 & 71.9 & 59.6 & 57.8 & 46.6 
            & 84.1 & 46.3 & 53.0 & 33.8 \\
            %& 61.0 & 38.8 & 43.2 & 26.6 \\
        \hspace{2mm} w/ DL4ND (ours)
            & -- & -- & \textbf{80.2} & \textbf{64.7} 
            & \textbf{68.4} & \textbf{54.1} 
            & -- & -- & \textbf{56.3} & \textbf{37.2} \\
            %& -- & -- & -- & -- \\
        \bottomrule
  \end{tabular}
  \label{app_table:synethic_erm}
\end{table*}

\begin{table}[t]
  \centering
  \setlength{\tabcolsep}{2pt}
  %\small
    \caption{Sensitivity analysis to warmup length on real-world noise top-1 accuracy on PACS~\citep{Li2017DeeperBA}. See \Cref{app_sec:extended_results} for discussion.}
    \vspace{-2mm}
   \begin{tabular}{lccc}
    \toprule
    Warmup Length   & ID & OOD &AVG \\
    \midrule
    w/o \methodAbbr & 96.3& 85.3 & 90.8 \\
    300 steps & 96.7 & 84.9  & 90.8\\
    600 steps & \textbf{97.3} & \textbf{88.8} & \textbf{93.1} \\
    1000 steps & 97.2 & 88.6 & 92.9 \\
    \bottomrule
  \end{tabular}
  \vspace{-2mm}
  \label{table:pacs_sensitivty}
\end{table}

\begin{table}
  \centering
  \setlength{\tabcolsep}{2pt}
  %\small
    \caption{Sensitivity analysis to warmup length on 20\% asymmetric noise on TerraIncognita~\citep{beery2018recognition}. See \Cref{app_sec:extended_results} for discussion.}
    \vspace{-2mm}
   \begin{tabular}{lccc}
    \toprule
    Warmup Length   & ID & OOD &AVG \\
    \midrule
    w/o \methodAbbr & \textbf{78.2}& 40.6 & 59.3 \\
    1000 steps & 77.4 & 41.2  & 59.3\\
    1200 steps & 77.1 & \textbf{43.5} & \textbf{60.3} \\
    1500 steps & 77.1 & 41.4 & 59.3 \\
    \bottomrule
  \end{tabular}
  \vspace{-2mm}
  \label{table:terra_sensitivty}
\end{table}
\Cref{table:pacs_sensitivty} and \Cref{table:terra_sensitivty} provides a sensitivity analysis on warmup length on PACS~\citep{Li2017DeeperBA} and TerraIncognita~\citep{beery2018recognition}, respectively.  Notably, the warmup length on either dataset is relatively short, but should be aided by a validation set to optimize its length.  That said, as the training dynamics during warmup does not change, setting the warmup length can be done in a single run (\ie, using a validation set to identify when performance gains start to decelerate). In addition, performance on PACS especially shows that the exact length choice can be relatively insensitive.

\subsection{Changing Sample Selection Ratios}
\label{sec:vlcs_sample_selection}

\begin{figure}[tbh]
  \centering
    \includegraphics[width=0.5\textwidth]{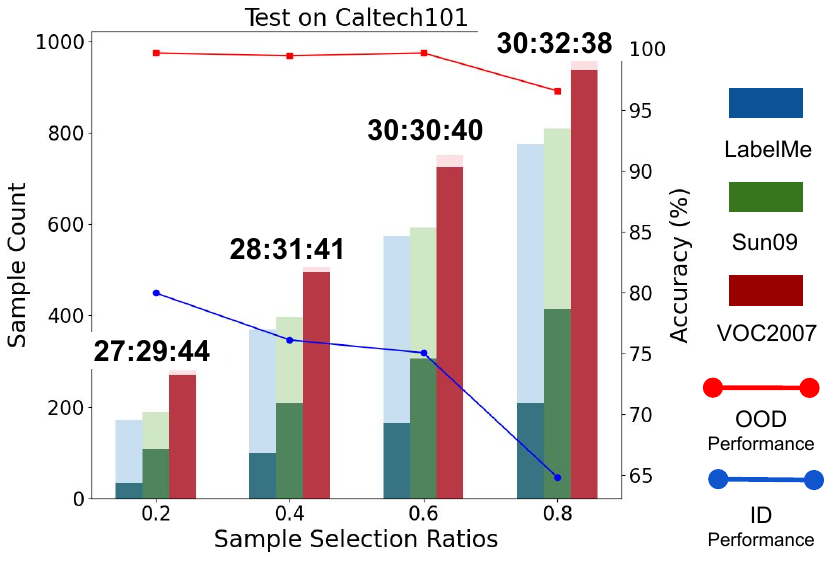}
  \caption{\textbf{Balance, clean sample ratios, and ID/OOD performance on VLCS ``Person" class.} Testing on Caltech101 with training on other domains. The x-axis shows sample selection ratios per class, with domain ratios above the bars. (\textit{Dark: clean samples; light: noisy.}) The decline in ID and OOD performance as balance increases suggests that a more balanced distribution does not always improve OOD accuracy, and increased noise harms both ID and OOD. See \Cref{sec:vlcs_sample_selection} for discussion.}
  \label{fig:vlcs_clean_size}
\end{figure}

\Cref{fig:vlcs_clean_size} shows the relationship between domain balance, clean sample count, and ID/OOD performance for the "person" class in VLCS. At lower selection ratios ($r$), the selected samples are cleaner but the distribution skews toward the cleaner VOC2007 domain, while higher ratios maintain balance but increase noise. The best results occur at $r = 0.2$, indicating that quality outweighs quantity for improved robustness.

\section{Comparison to DA+LNL Methods}
\label{sec:dalnl}

As discussed in our related work section, where are several methods that explore the intersection of Domain Adaption (DA) and LNL (\eg,~\citet{10.1609/aaai.v33i01.33014951,9921307,9488210,10.1109/TMM.2022.3205457,10.1145/3581783.3612296,Yin_Feng_Yan_Song_Peng_Wang_2025}).  In DA their goal is to improve performance on a target domain when given a labeled source domain and unlabeled samples from a target domain.   However, in DG, which we study, the goal is to generalize to unseen distributions, thereby requiring different methods.  Additionally, they are evaluated on how well they perform on the target domain, which is similar to DG methods that evaluate performance only on unseen domains (ignoring source domain performance).  As we show, this often results in performing well only on the unseen domains, thereby limiting its applications when both ID and OOD performance is desired (\eg, MIRO in \Cref{table:real-world_combos}(a) obtains a 4\% boost to OOD performance on CHAMMI, but at a cost to 12\% ID performance).

Further, when we consider adapting the ideas of these methods to our task, we find that many of them are well represented in the LNL literature.  For example, \citet{Yin_Feng_Yan_Song_Peng_Wang_2025,9488210} separates samples into a clean and noisy set, and then aims to use the clean set to correct the noisy labels, which is also the general approach taken by relabeling methods like DISC~\citep{li2023disc} and UNICON~\citep{karim2022unicon}.  Similarly, \citet{10.1145/3581783.3612296,10.1109/TMM.2022.3205457} aims to estimate label uncertainty and then re-weight samples based on this uncertainty, which is also the general approach of LNL method PLM~\citep{zhao2024estimating}.  As seen in \Cref{table:real-world_combos}(b), DISC, UNICON, and PLM underperform our \methodAbbrspace approach by up to 18\%.

The reason these methods do not transfer well is due, at least in part, to the differences in DA and DG, as there tends to only be a transfer between a single source domain and a single target domain.  Specifically, if we consider the example in \Cref{fig:intro} of our paper, the main problem we raise is that in NAG we must learn to separate label noise from domain shifts.  However, in DA there is no need to solve this problem as there is only a single source domain with noise.  Thus, they are more akin to the traditional LNL task that aims to identify noise in a single domain, which is likely why similar ideas appear between them.  We simulate this type of setting by using per-domain sampling in \Cref{table:naive_combo}, effectively converting the problem into a set of single-domain noise detection tasks. However, this underperforms our approach by 2\% on VLCS. This not only highlights the differences between the tasks but also illustrates the benefits of our cross-domain comparisons.

\section{Real-world Dataset's Noise}
\label{app:data_noise}

\begin{figure}[t]
  \centering
    %\vspace{-2mm}
    \includegraphics[width=0.5\textwidth]{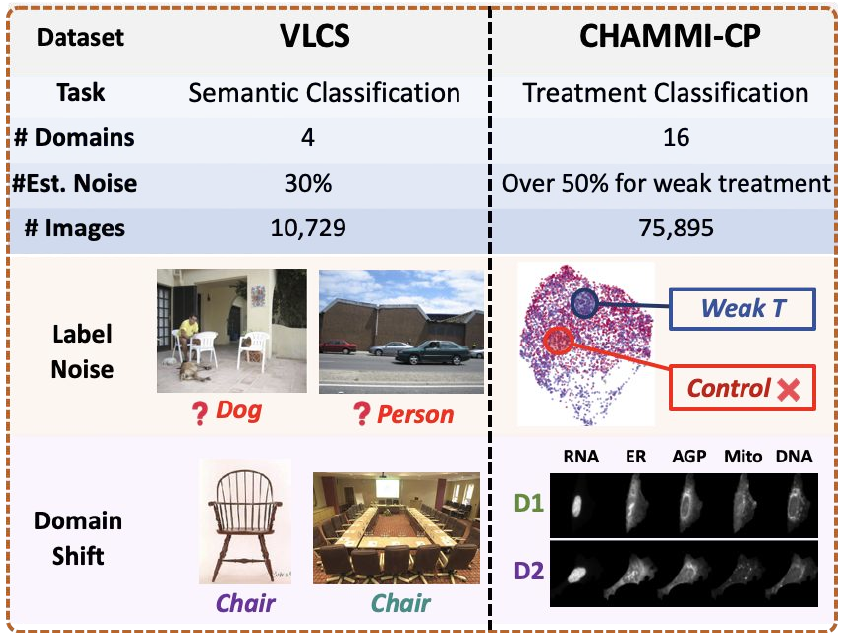}
  \caption{\textbf{Real-world datasets with in-domain noise and multi-domain distribution}. \textbf{VLCS} (web/user data)~\citep{fang2013unbiased}, and \textbf{CHAMMI-CP} (biomedical images)~\citep{chen2024chammi}. VLCS faces label noise from poor annotations and domain shifts from varying data sources, while CHAMMI-CP deals with ambiguous features and varying experimental environments. }
  \label{fig:fig2}
\end{figure}

\Cref{fig:fig2} reports statistics on the two datasets used most extensively in our experiments, VLCS~\citep{fang2013unbiased} and CHAMMI-CP~\citep{chen2024chammi}.  Generally speaking, CHAMMI-CP is larger, with more domains and is very noisy.  However, while there are many domains, these represent what can be thought of as reproduction experiments. \Ie, the variations that are shown are due to technical variations that arise when reproducing an experiment within the same laboratory setting (\ie, same equipment).  As such, their domains are more similar than those in VLCS.  

\Cref{tab:vlcs_summary} provides more detailed statistics about the source of the noise in VLCS.  Notably, the various domains have different degrees of noise, \eg, Caltech~\citep{fei2004learning-caltech} is relatively clean, whereas LabelMe~\citep{russell2008labelme} and SUN09~\citep{choi2010exploiting-sun09} are much noisier.  The main source of the noise is due to unlabeled categories.  For example, person images that do not label cars or chairs are common.  As such, while it is the person images being labeled as noisy, the actual issue is that car is not also annotated.  This makes this noise type similar to the asymmetric noise used in our synthetic noise experiments in \Cref{sec:exp}.  Our inspection of PACS~\citep{Li2017DeeperBA} finds it contains the same type of noise as VLCS.  

\begin{table}[ht]
\centering
\caption{VLCS Dataset Overview (Total Samples, Noisy Samples) over its datasets Caltech~\citep{fei2004learning-caltech}, LabelMe~\citep{russell2008labelme}, SUN09~\citep{choi2010exploiting-sun09}, and VOC2007~\citep{everingham2010pascal-voc2007}.}
\begin{tabular}{l|l|c|c}
\hline
\textbf{Domain}   & \textbf{Category} & \textbf{Total Samples} & \textbf{Noisy Samples} \\ \hline
\multirow{5}{*}{Caltech} 
    & Bird    & 237  & 1 \\ 
    &         &      & (with person) \\ \cline{2-4}
    & Car     & 123  & 0 \\ 
    &         &      & (black \& white car imgs) \\ \cline{2-4}
    & Chair   & 118  & 0 \\ \cline{2-4}
    & Dog     & 67   & 0 \\ 
    &         &      & (only black and white dog) \\ \cline{2-4}
    & Person  & 870  & 0 \\ 
    &         &      & (profile photos with redundancy) \\ \hline
\multirow{5}{*}{LabelMe} 
    & Bird    & 80   & 20 \\ \cline{2-4}
    & Car     & 1209 & 559 \\ 
    &         &      & (background: building, road, mountains; \\ 
    &         &      & small \& incomplete cars, unclear night imgs [OOD]) \\ \cline{2-4}
    & Chair   & 89   & 61 \\ 
    &         &      & (over half have cars, person) \\ \cline{2-4}
    & Dog     & 43   & 25 \\ 
    &         &      & (with person, cars) \\ \cline{2-4}
    & Person  & 1238 & 924 \\ 
    &         &      & (over 80\% noisy images have cars, \\ 
    &         &      & street photos are similar to car and chair categories, \\ 
    &         &      & small person figures) \\ \hline
\multirow{5}{*}{SUN09} 
    & Bird    & 21   & 12 \\ 
    &         &      & (background, 1 person and dog) \\ \cline{2-4}
    & Car     & 933  & 548 \\ 
    &         &      & (street view, buildings, person) \\ \cline{2-4}
    & Chair   & 1036 & 186 \\ 
    &         &      & (mostly person, very few car interior) \\ \cline{2-4}
    & Dog     & 31   & 25 \\ 
    &         &      & ($\sim$20 noisy images with person) \\ \cline{2-4}
    & Person  & 1265 & 631 \\ 
    &         &      & (very small person figures) \\ \hline
\multirow{5}{*}{VOC2007} 
    & Bird    & 330  & 29 \\ 
    &         &      & (mostly human, a few cars, one small bird) \\ \cline{2-4}
    & Car     & 699  & 133 \\ 
    &         &      & (mostly person, $\sim$5 don't have cars) \\ \cline{2-4}
    & Chair   & 428  & 145 \\ 
    &         &      & (mostly person, some cars, very few missing chair) \\ \cline{2-4}
    & Dog     & 420  & 111 \\ 
    &         &      & (mostly human, a few cars) \\ \cline{2-4}
    & Person  & 1499 & 61 \\ 
    &         &      & (mostly cars, some don't have person) \\ \hline
\end{tabular}
\label{tab:vlcs_summary}
\end{table}

\subsection{Synthesized Noise}
\label{app:exp_syn_noise}

Our synthetic experiments use asymmetric noise, which flips labels between commonly confused categories.  These confusions can, but are not required to go both directions, and are manually selected. We provide our category confusion information in \Cref{tab:asym_rotated} for  RotatedMNIST~\citep{ghifary2015domain}, \Cref{tab:asym_office} for OfficeHome~\citep{venkateswara2017deep}, \Cref{tab:asym_terra} for TerraIncognita~\citep{beery2018recognition}, and \Cref{tab:asym_domain} for DomainNet~\citep{peng2019moment}.

\begin{table}
    \centering
    \caption{Asymmetric Noise Pairs for RotatedMNIST~\citep{ghifary2015domain}.}
    \begin{tabular}{c c}
        \toprule
        Index A & Index B \\
        \midrule
        0 & 6 \\
        1 & 7 \\
        3 & 5 \\
        4 & 9 \\
        5 & 3 \\
        6 & 0 \\
        7 & 1 \\
        9 & 4 \\
        \bottomrule
    \end{tabular}
    \label{tab:asym_rotated}
\end{table}

\begin{table}
    \centering
    \caption{Asymmetric Noise Pairs for OfficeHome~\citep{venkateswara2017deep}.}
    \begin{tabular}{c c c c}
        \toprule
        Index A & Class A & Index B & Class B \\
        \midrule
        16 & Pencil & 6 & Pen \\
        14 & Keyboard & 42 & Laptop \\
        15 & Mouse & 60 & Monitor \\
        10 & Backpack & 39 & Clipboards \\
        1 & Calculator & 34 & Notebook \\
        47 & Bottle & 63 & Soda \\
        13 & Flowers & 21 & Candles \\
        3 & Flipflops & 54 & Sneakers \\
        9 & TV & 60 & Monitor \\
        8 & Speaker & 53 & Radio \\
        4 & Kettle & 52 & Pan \\
        19 & Webcam & 42 & Laptop \\
        5 & Mop & 56 & Bucket \\
        24 & Knives & 32 & Fork \\
        12 & Desk Lamp & 33 & Lamp Shade \\
        18 & Spoon & 32 & Fork \\
        17 & Scissors & 27 & Screwdriver \\
        50 & Hammer & 22 & Drill \\
        48 & Computer & 60 & Monitor \\
        23 & Folder & 34 & Notebook \\
        26 & Post-it Notes & 61 & Paper Clip \\
        58 & File Cabinet & 36 & Shelf \\
        44 & Push Pin & 26 & Post-it Notes \\
        45 & Sink & 62 & Refrigerator \\
        49 & Fan & 33 & Lamp Shade \\
        25 & Mug & 47 & Bottle \\
        57 & Couch & 30 & Chair \\
        \bottomrule
    \end{tabular}
    \label{tab:asym_office}
\end{table}

\begin{table}
    \centering
    \caption{Asymmetric Noise Pairs for TerraIncognita~\citep{beery2018recognition}.}
    \begin{tabular}{c c c c}
        \toprule
        Index A & Class A & Index B & Class B \\
        \midrule
        0 & Bird & 9 & Squirrel \\
        1 & Bobcat & 3 & Coyote \\
        2 & Cat & 4 & Dog \\
        3 & Coyote & 8 & Raccoon \\
        4 & Dog & 2 & Cat \\
        5 & Empty & 0 & Bird \\
        6 & Opossum & 8 & Raccoon \\
        7 & Rabbit & 9 & Squirrel \\
        8 & Raccoon & 6 & Opossum \\
        9 & Squirrel & 7 & Rabbit \\
        \bottomrule
    \end{tabular}
    \label{tab:asym_terra}
\end{table}

\begin{table}
\centering
\small
\caption{Asymmetric Noise Pairs for DomainNet~\citep{peng2019moment}.}
\begin{tabular}{|cc||cc||cc||cc||cc||cc|}
\toprule
Key & Value & Key & Value & Key & Value & Key & Value & Key & Value & Key & Value \\
\midrule
0 & 308 & 1 & 208 & 2 & 28 & 3 & 135 & 4 & 5 & 5 & 0 \\
6 & 0 & 7 & 324 & 8 & 324 & 9 & 208 & 10 & 288 & 11 & 324 \\
12 & 208 & 13 & 285 & 14 & 208 & 15 & 16 & 16 & 17 & 17 & 282 \\
18 & 19 & 19 & 327 & 20 & 309 & 21 & 208 & 22 & 327 & 23 & 208 \\
24 & 288 & 25 & 135 & 26 & 27 & 27 & 28 & 28 & 208 & 29 & 208 \\
30 & 327 & 31 & 98 & 32 & 33 & 33 & 144 & 34 & 35 & 35 & 308 \\
36 & 282 & 37 & 38 & 38 & 327 & 39 & 208 & 40 & 208 & 41 & 42 \\
42 & 208 & 43 & 44 & 44 & 308 & 45 & 46 & 46 & 331 & 47 & 324 \\
48 & 91 & 49 & 90 & 50 & 327 & 51 & 324 & 52 & 53 & 53 & 324 \\
54 & 327 & 55 & 331 & 56 & 282 & 57 & 151 & 58 & 334 & 59 & 324 \\
60 & 324 & 61 & 208 & 62 & 175 & 63 & 64 & 64 & 327 & 65 & 208 \\
66 & 67 & 67 & 68 & 68 & 208 & 69 & 208 & 70 & 138 & 71 & 331 \\
72 & 324 & 73 & 175 & 74 & 53 & 75 & 254 & 76 & 338 & 77 & 276 \\
78 & 91 & 79 & 208 & 80 & 282 & 81 & 208 & 82 & 282 & 83 & 319 \\
84 & 85 & 85 & 208 & 86 & 310 & 87 & 324 & 88 & 208 & 89 & 90 \\
90 & 91 & 91 & 208 & 92 & 323 & 93 & 285 & 94 & 95 & 95 & 261 \\
96 & 276 & 97 & 98 & 98 & 324 & 99 & 282 & 100 & 288 & 101 & 102 \\
102 & 103 & 103 & 327 & 104 & 110 & 105 & 288 & 106 & 107 & 107 & 282 \\
108 & 276 & 109 & 110 & 110 & 324 & 111 & 110 & 112 & 288 & 113 & 114 \\
114 & 157 & 115 & 208 & 116 & 327 & 117 & 98 & 118 & 327 & 119 & 208 \\
120 & 208 & 121 & 110 & 122 & 324 & 123 & 208 & 124 & 125 & 125 & 208 \\
126 & 208 & 127 & 324 & 128 & 129 & 129 & 208 & 130 & 327 & 131 & 208 \\
132 & 208 & 133 & 28 & 134 & 135 & 135 & 136 & 136 & 324 & 137 & 138 \\
138 & 35 & 139 & 282 & 140 & 324 & 141 & 208 & 142 & 208 & 143 & 282 \\
144 & 324 & 145 & 146 & 146 & 282 & 147 & 148 & 148 & 208 & 149 & 208 \\
150 & 151 & 151 & 98 & 152 & 153 & 153 & 308 & 154 & 208 & 155 & 341 \\
156 & 157 & 157 & 208 & 158 & 324 & 159 & 208 & 160 & 208 & 161 & 98 \\
162 & 163 & 163 & 208 & 164 & 282 & 165 & 308 & 166 & 230 & 167 & 1 \\
168 & 285 & 169 & 208 & 170 & 171 & 171 & 208 & 172 & 208 & 173 & 208 \\
174 & 175 & 175 & 208 & 176 & 282 & 177 & 178 & 178 & 110 & 179 & 246 \\
180 & 208 & 181 & 282 & 182 & 324 & 183 & 282 & 184 & 208 & 185 & 324 \\
186 & 324 & 187 & 188 & 188 & 282 & 189 & 190 & 190 & 324 & 191 & 282 \\
192 & 193 & 193 & 135 & 194 & 35 & 195 & 28 & 196 & 282 & 197 & 307 \\
198 & 178 & 199 & 208 & 200 & 208 & 201 & 28 & 202 & 324 & 203 & 282 \\
204 & 208 & 205 & 206 & 206 & 282 & 207 & 208 & 208 & 91 & 209 & 324 \\
210 & 211 & 211 & 212 & 212 & 213 & 213 & 288 & 214 & 208 & 215 & 216 \\
216 & 282 & 217 & 246 & 218 & 335 & 219 & 276 & 220 & 282 & 221 & 222 \\
222 & 208 & 223 & 327 & 224 & 110 & 225 & 285 & 226 & 208 & 227 & 228 \\
228 & 208 & 229 & 324 & 230 & 327 & 231 & 232 & 232 & 208 & 233 & 282 \\
234 & 282 & 235 & 324 & 236 & 327 & 237 & 208 & 238 & 285 & 239 & 240 \\
240 & 331 & 241 & 285 & 242 & 324 & 243 & 208 & 244 & 309 & 245 & 107 \\
246 & 247 & 247 & 248 & 248 & 324 & 249 & 321 & 250 & 251 & 251 & 288 \\
252 & 135 & 253 & 254 & 254 & 327 & 255 & 208 & 256 & 208 & 257 & 341 \\
258 & 208 & 259 & 135 & 260 & 261 & 261 & 262 & 262 & 208 & 263 & 213 \\
264 & 208 & 265 & 327 & 266 & 208 & 267 & 268 & 268 & 269 & 269 & 208 \\
270 & 309 & 271 & 208 & 272 & 273 & 273 & 135 & 274 & 208 & 275 & 276 \\
276 & 277 & 277 & 324 & 278 & 279 & 279 & 208 & 280 & 281 & 281 & 282 \\
282 & 282 & 283 & 208 & 284 & 285 & 285 & 98 & 286 & 282 & 287 & 208 \\
288 & 310 & 289 & 324 & 290 & 282 & 291 & 309 & 292 & 208 & 293 & 294 \\
294 & 208 & 295 & 324 & 296 & 327 & 297 & 208 & 298 & 208 & 299 & 324 \\
300 & 208 & 301 & 285 & 302 & 324 & 303 & 282 & 304 & 282 & 305 & 282 \\
306 & 307 & 307 & 308 & 308 & 282 & 309 & 282 & 310 & 341 & 311 & 208 \\
312 & 313 & 313 & 331 & 314 & 282 & 315 & 282 & 316 & 282 & 317 & 282 \\
318 & 282 & 319 & 327 & 320 & 327 & 321 & 282 & 322 & 208 & 323 & 324 \\
324 & 325 & 325 & 324 & 326 & 327 & 327 & 282 & 328 & 329 & 329 & 282 \\
330 & 282 & 331 & 332 & 332 & 324 & 333 & 282 & 334 & 335 & 335 & 208 \\
336 & 337 & 337 & 338 & 338 & 208 & 339 & 340 & 340 & 341 & 341 & 342 \\
342 & 208 & 343 & 344 & 344 & 282 & \multicolumn{2}{c||}{} & \multicolumn{2}{c||}{} & \multicolumn{2}{c|}{}\\
\bottomrule
\end{tabular}
\label{tab:asym_domain}
\end{table}

\section{Impelmentation details}
\label{app:implementation_details}

We incorporate the implementation of the ERM++~\footnote{https://github.com/piotr-teterwak/erm\_plusplus}~\citep{teterwak2023erm++}, DISC~\footnote{https://github.com/JackYFL/DISC}~\citep{li2023disc}, UNICON~\footnote{https://github.com/nazmul-karim170/UNICON-Noisy-Label}~\citep{karim2022unicon}, ELR~\footnote{https://github.com/shengliu66/ELR}~\citep{liu2020early}, SAGM~\footnote{https://github.com/Wang-pengfei/SAGM}~\citep{wang2023sharpness},
MIRO~\footnote{https://github.com/kakaobrain/miro}~\citep{cha2022miro},
VREx~\footnote{https://github.com/facebookresearch/DomainBed}~\citep{krueger2021out},
Fishr~\footnote{https://github.com/alexrame/fishr}~\citep{rame2022fishr},
DISAM~\footnote{https://github.com/MediaBrain-SJTU/DISAM}~\citep{zhang2024domain},
PLM~\footnote{https://github.com/RyanZhaoIc/PLM/tree/main}~\citep{zhao2024estimating},
into our codebase. Each training batch includes samples from all training domains, with a batch size of 128. For relatively small datasets VLCS~\citep{fang2013unbiased} and CHAMMI-CP~\citep{chen2024chammi}, experiments are run on a single NVIDIA RTX A6000 (48GB RAM) and three Intel(R) Xeon(R) Gold 6226R CPU @ 2.90GHz for 5000 steps. Below we give a short description of the methods in prior work used in our experiments.

\subsection{Learning with Noisy Labels Methods}

\noindent\textbf{ELR}~\citep{liu2020early} is based on the observation that deep neural networks initially fit the training data with clean labels during an "early learning" phase before eventually memorizing examples with false labels. It employs semi-supervised learning to generate target probabilities and introduces a regularization term to prevent memorizing false labels, guiding the model toward these target probabilities.

\noindent\textbf{UNICON}~\citep{karim2022unicon} partitions the training set into clean and noisy subsets using uniform selection at each iteration. It estimates the clean label probability via Jensen-Shannon divergence (JSD) based on prediction and one-hot label distribution disagreement. After partitioning, UNICON applies semi-supervised learning (SSL) with a contrastive loss across two identical networks, repeating this process until convergence.

\noindent\textbf{DISC}~\citep{li2023disc} identifies clean samples based on predictions from weak and strong augmentations, using a dynamic confidence threshold determined by each instance's memorization strength from previous epochs. This approach classifies instances into clean, hard, and purified subsets. Different regularization strategies are then applied to each subgroup.

\noindent\textbf{PLM}~\citep{zhao2024estimating} is a classifier-consistent method, which estimates the noisy class posterior with noise transition matrix to correct the label in the training. PLM crops instances to generate part-level labels, which are then modeled with a novel single-to-multiple transition matrix to capture the relationship between noisy and part-level labels. 

\subsubsection{Domain Generalization Methods}

\noindent\textbf{ERM}~\citep{gulrajani2020search-erm} is the simplest baseline method, where models are simply trained on the multiple sources. 

\noindent\textbf{ERM++}~\citep{teterwak2023erm++} is an enhanced baseline that tunes training components to mitigate overfitting and boost generalization performance. It employs a two-stage training pipeline, explores strong initialization strategies, and investigates regularization techniques such as Model Parameter Averaging, Warm Start, Unfreezing BatchNorm, and Attention Tuning.

\noindent\textbf{MIRO}~\citep{cha2022miro}  is a DG framework that guides learning by maximizing mutual information between oracle representations, approximated with pre-trained models. The objective function combines empirical risk and a variational bound of the mutual information, effectively enhancing generalization ability.

\noindent\textbf{SWAD}~\citep{cha2021swad} improves DG performance by finding flat minima in the loss landscape. It extends SWA with a dense sampling strategy and an overfit-aware sampling schedule, resulting in flatter minima and better generalization across domains while being robust to model selection and overfitting issues.

\noindent\textbf{Fishr}~\citep{rame2022fishr} is a regularization method that enforces domain invariance by matching the domain-level variances of gradients across training domains. By aligning the domain-level loss landscapes around the final weights, Fisher effectively improves out-of-distribution generalization.

\noindent\textbf{VREx}~\citep{krueger2021out} is a method that aims to reduce differences in risk across training domains to enhance robustness against extreme distributional shifts,with the assumption that domain shift is a type of variation.  It proposes a penalty on the variance of training risks.

\noindent\textbf{SAGM}~\citep{wang2023sharpness} boosts generalization across domains by finding flat regions with low loss. It minimizes the empirical risk loss, perturbed loss, and surrogate gap while performing gradient matching, efficiently guiding the model towards flat, low-loss regions, resulting in better generalization compared to SAM-like methods.

\noindent\textbf{DISAM}~\citep{zhang2024domain} addresses domain-level convergence consistency in sharpness-aware minimization (SAM) by introducing a constraint to minimize variance in domain losses. This approach prevents excessive or insufficient perturbations in domains that are less or more well-optimized.

\subsection{Details on DG+LNL integration}

Algorithm~\ref{alg:ermplus_elr}, ~\ref{alg:miro_elr}, ~\ref{alg:swad_elr}, ~\ref{alg:miro_swad_elr}, ~\ref{alg:miro_unicon}, ~\ref{alg:miro_swad_unicon} show the detail of the integrated methods.

\begin{algorithm}
\caption{ERM++ + ELR Algorithm.}
\label{alg:ermplus_elr}
\Input{Sample inputs $X=\{x_i\}_{i=1}^n$, noisy labels $\widetilde{Y}=\{\widetilde{y_i}\}_{i=1}^n$, ELR temporal ensembling momentum $\beta$, regularization parameter $\lambda$, neural network with trainable parameters $f_\theta$}
\Output{Neural network with updated parameters $f_{\theta'}$}
\For{$step\gets1$ \KwTo $training\_steps$}{
    \For{minibatch $B$}{
        \For{$i$ in $B$}{
            $p_i = f_\theta(x_i) $ \tcp*[l]{Model prediction.}
            $t_i = \beta*t_i + (1-\beta)*p_i $ \tcp*[l]{Temporal ensembling.}
        }
        loss = $ -\frac{1}{|B|} \Sigma_{|B|} cross\_entropy (p_i, y_i) + \frac{\lambda}{|B|}\Sigma_{|B|} log(1-<p_i, t_i>)$ \tcp*[l]{ELR loss: cross entropy loss and regularization loss.}
        Update $f_\theta$.
    }
    $f_{\theta'}$ = Update $f_{\theta}$ with ERM++ parameter averaging.
}
\end{algorithm}

\begin{algorithm}
\caption{MIRO + ELR Algorithm.}
\label{alg:miro_elr}
\Input{Sample inputs $X=\{x_i\}_{i=1}^n$, noisy labels $\widetilde{Y}=\{\widetilde{y_i}\}_{i=1}^n$, ELR temporal ensembling momentum $\beta$, ELR regularization parameter $\lambda1$, MIRO regularization parameter $\lambda2$, MIRO mean encoder $\mu$, MIRO variance encode $\sigma$, feature extractor with trainable parameters $f_\theta$, pretrained feature extractor with parameters $f_{\theta_0}$}
\Output{Neural network with updated parameters $f_{\theta'}$}
\For{$step\gets1$ \KwTo $training\_steps$}{
    \For{minibatch $B$}{
        \For{$i$ in $B$}{
            $p_i = f_\theta(x_i) $ \tcp*[l]{feature extractor output.}
            $p_i^{0} = f_{\theta_0}(x_i) $ \tcp*[l]{Pretrained feature extractor output.}
            $t_i = \beta*t_i + (1-\beta)*p_i $ \tcp*[l]{Temporal ensembling.}
        }
        loss = $ -\frac{1}{|B|} \Sigma_{|B|} cross\_entropy (p_i, y_i)$ \tcp*[l]{Cross entropy loss.}
        loss +=$ \frac{\lambda1}{|B|}\Sigma_{|B|} log(1-<p_i, t_i>)$ \tcp*[l]{ELR loss with regularization term.}
        loss += $\frac{\lambda2}{|B|}\Sigma_{|B|} (log(|\sigma(p_i)|) + ||p_i^{0}-\mu(p_i)||^2_{\sigma(p_i)^{-1}})$ \tcp*[l]{MIRO loss with regularization term.}
        Update $f_\theta$.
    }
    $f_{\theta'}$ = Updated $f_{\theta}$.
}
\end{algorithm}

\begin{algorithm}
\caption{SWAD + ELR Algorithm.}
\label{alg:swad_elr}
\Input{Sample inputs $X=\{x_i\}_{i=1}^n$, noisy labels $\widetilde{Y}=\{\widetilde{y_i}\}_{i=1}^n$, ELR temporal ensembling momentum $\beta$, ELR regularization parameter $\lambda$, neural network with trainable parameters $f_\theta$}
\Output{Neural network with updated parameters $f_{\theta'}$}
\For{$step\gets1$ \KwTo $training\_steps$}{
    \For{minibatch $B$}{
        \For{$i$ in $B$}{
            $p_i = f_\theta(x_i) $ \tcp*[l]{Model prediction.}
            $t_i = \beta*t_i + (1-\beta)*p_i $ \tcp*[l]{Temporal ensembling.}
        }
        loss = $ -\frac{1}{|B|} \Sigma_{|B|} cross\_entropy (p_i, y_i) + \frac{\lambda}{|B|}\Sigma_{|B|} log(1-<p_i, t_i>)$ \tcp*[l]{ELR loss: cross entropy loss and regularization loss.}
        Update $f_\theta$.
        Decide the start $step_s$ and end $step_e$ iteration for averaging in SWAD.
    }
    $f_{\theta'}$ = $\frac{1}{step_e - step_s+1} \Sigma f_\theta$ \tcp*[l]{SWAD parameter averaging.}
}
\end{algorithm}

\begin{algorithm}
\caption{MIRO + SWAD + ELR Algorithm.}
\label{alg:miro_swad_elr}
\Input{Sample inputs $X=\{x_i\}_{i=1}^n$, noisy labels $\widetilde{Y}=\{\widetilde{y_i}\}_{i=1}^n$, ELR temporal ensembling momentum $\beta$, ELR regularization parameter $\lambda1$, MIRO regularization parameter $\lambda2$, MIRO mean encoder $\mu$, MIRO variance encode $\sigma$, feature extractor with trainable parameters $f_\theta$, pretrained feature extractor with parameters $f_{\theta_0}$}
\Output{Neural network with updated parameters $f_{\theta'}$}
\For{$step\gets1$ \KwTo $training\_steps$}{
    \For{minibatch $B$}{
        \For{$i$ in $B$}{
            $p_i = f_\theta(x_i) $ \tcp*[l]{feature extractor output.}
            $p_i^{0} = f_{\theta_0}(x_i) $ \tcp*[l]{Pretrained feature extractor output.}
            $t_i = \beta*t_i + (1-\beta)*p_i $ \tcp*[l]{Temporal ensembling.}
        }
        loss = $ -\frac{1}{|B|} \Sigma_{|B|} cross\_entropy (p_i, y_i)$ \tcp*[l]{Cross entropy loss.}
        loss +=$ \frac{\lambda1}{|B|}\Sigma_{|B|} log(1-<p_i, t_i>)$ \tcp*[l]{ELR loss with regularization term.}
        loss += $\frac{\lambda2}{|B|}\Sigma_{|B|} (log(|\sigma(p_i)|) + ||p_i^{0}-\mu(p_i)||^2_{\sigma(p_i)^{-1}})$ \tcp*[l]{MIRO loss with regularization term.}
        Update $f_\theta$.
        Decide the start $step_s$ and end $step_e$ iteration for averaging in SWAD.
    }
     $f_{\theta'}$ = $\frac{1}{step_e - step_s+1} \Sigma f_\theta$ \tcp*[l]{SWAD parameter averaging.}
}
\end{algorithm}

\begin{algorithm}
\caption{MIRO + UNICON Algorithm.}
\label{alg:miro_unicon}
\Input{Sample inputs $X=\{x_i\}_{i=1}^n$, noisy labels $\widetilde{Y}=\{\widetilde{y_i}\}_{i=1}^n$, MIRO regularization parameter $\lambda2$, MIRO mean encoder $\mu$, MIRO variance encode $\sigma$, feature extractor-1 with trainable parameters $f1_\theta$, feature extractor-2 with trainable parameters $f2_\theta$, pretrained feature extractor with parameters $f_{\theta_0}$, UNICON sharpening temperature $T$, UNICON unsupervised loss coefficient $\lambda_u$, UNICON contrastive loss coefficient $\lambda_c$, , UNICON regularization loss coefficient $\lambda_r$.}
\Output{Neural network with updated parameters $f1_{\theta'}$ and $f2_{\theta'}$}
\For{$step\gets1$ \KwTo $training\_steps$}{
    $D_{clean}, D_{noisy} = UNICON-Selection (X=\{x_i\}_{i=1}^n, f1_\theta, f2_\theta)$, \tcp*[l]{UNICON clean-noisy sample selection.}
    \For{clean minibatch $B_{clean}$}{
        \For{noisy minibatch $B_{noisy}$}{
             \For{$i$ in $B = B_{clean} \bigcup B_{noisy}$}{
                $p1_i = f1_\theta(x_i) $ \tcp*[l]{feature extractor-1 output.}
                $p2_i = f2_\theta(x_i) $ \tcp*[l]{feature extractor-2 output.}
                $p_i^{0} = f_{\theta_0}(x_i) $ \tcp*[l]{Pretrained feature extractor output.}
            }
            $loss_1 =  -\frac{1}{|B|} \Sigma_{|B|} cross\_entropy (p1_i, y_i)$ \tcp*[l]{Cross entropy loss for feature extractor-1.}
            $loss_1 += \frac{\lambda2}{|B|}\Sigma_{|B|} (log(|\sigma(p1_i)|) + ||p_i^{0}-\mu(p1_i)||^2_{\sigma(p1_i)^{-1}})$ \tcp*[l]{MIRO loss with regularization term for feature extractor-1.}
            $loss_2 =  -\frac{1}{|B|} \Sigma_{|B|} cross\_entropy (p2_i, y_i)$ \tcp*[l]{Cross entropy loss for feature extractor-2.}
            $loss_2 += \frac{\lambda2}{|B|}\Sigma_{|B|} (log(|\sigma(p2_i)|) + ||p_i^{0}-\mu(p2_i)||^2_{\sigma(p2_i)^{-1}})$ \tcp*[l]{MIRO loss with regularization term for feature extractor-2.}
            $X_{clean|B|}^{weak}$ = weak-augmentation($B_{clean}$)\\
            $X_{noisy|B|}^{weak}$ = weak-augmentation($B_{noisy}$)\\
            $X_{clean|B|}^{strong}$ = strong-augmentation($B_{clean}$)\\
            $X_{noisy|B|}^{strong}$ = strong-augmentation($B_{noisy}$)\\
            Get labeled set with UNICON label refinement on clean batch.\\
            Get unlabeled set with UNICON pseudo label on noisy batch.\\
            $L_{u1}, L_{u2}$ = MixMatch on labeled and unlabeled sets \tcp*[l]{UNICON unsupervised loss for feature extractor-1 and extractor-2.}
            Get $L_{c1}, L_{c2}$  \tcp*[l]{UNICON contrastive loss for feature extractor-1 and extractor-2.}
            Get $L_{r1}, L_{r2}$  \tcp*[l]{UNICON regularization loss for feature extractor-1 and extractor-2.}
            $loss_1 += \lambda_u* L_{u1} + \lambda_c* L_{c1} + \lambda_r* L_{r1}$ \tcp*[l]{Update UNICON loss for feature extractor-1.}
            $loss_2 += \lambda_u* L_{u2} + \lambda_c* L_{c2} + \lambda_r* L_{r2}$ \tcp*[l]{Update UNICON loss for feature extractor-2.}
            Update $f1_\theta$ and $f2_\theta$.
        }
    }
    $f1_{\theta'}$ = Updated $f1_{\theta}$, $f2_{\theta'}$ = Updated $f2_{\theta}$.
}
\end{algorithm}

\begin{algorithm}
\caption{MIRO + SWAD + UNICON Algorithm.}
\label{alg:miro_swad_unicon}
\Input{Sample inputs $X=\{x_i\}_{i=1}^n$, noisy labels $\widetilde{Y}=\{\widetilde{y_i}\}_{i=1}^n$, MIRO regularization parameter $\lambda2$, MIRO mean encoder $\mu$, MIRO variance encode $\sigma$, feature extractor-1 with trainable parameters $f1_\theta$, feature extractor-2 with trainable parameters $f2_\theta$, pretrained feature extractor with parameters $f_{\theta_0}$, UNICON sharpening temperature $T$, UNICON unsupervised loss coefficient $\lambda_u$, UNICON contrastive loss coefficient $\lambda_c$, , UNICON regularization loss coefficient $\lambda_r$.}
\Output{Neural network with updated parameters $f1_{\theta'}$ and $f2_{\theta'}$}
\For{$step\gets1$ \KwTo $training\_steps$}{
    $D_{clean}, D_{noisy} = UNICON-Selection (X=\{x_i\}_{i=1}^n, f1_\theta, f2_\theta)$, \tcp*[l]{UNICON clean-noisy sample selection.}
    \For{clean minibatch $B_{clean}$}{
        \For{noisy minibatch $B_{noisy}$}{
             \For{$i$ in $B = B_{clean} \bigcup B_{noisy}$}{
                $p1_i = f1_\theta(x_i) $ \tcp*[l]{feature extractor-1 output.}
                $p2_i = f2_\theta(x_i) $ \tcp*[l]{feature extractor-2 output.}
                $p_i^{0} = f_{\theta_0}(x_i) $ \tcp*[l]{Pretrained feature extractor output.}
            }
            $loss_1 =  -\frac{1}{|B|} \Sigma_{|B|} cross\_entropy (p1_i, y_i)$ \tcp*[l]{Cross entropy loss for feature extractor-1.}
            $loss_1 += \frac{\lambda2}{|B|}\Sigma_{|B|} (log(|\sigma(p1_i)|) + ||p_i^{0}-\mu(p1_i)||^2_{\sigma(p1_i)^{-1}})$ \tcp*[l]{MIRO loss with regularization term for feature extractor-1.}
            $loss_2 =  -\frac{1}{|B|} \Sigma_{|B|} cross\_entropy (p2_i, y_i)$ \tcp*[l]{Cross entropy loss for feature extractor-2.}
            $loss_2 += \frac{\lambda2}{|B|}\Sigma_{|B|} (log(|\sigma(p2_i)|) + ||p_i^{0}-\mu(p2_i)||^2_{\sigma(p2_i)^{-1}})$ \tcp*[l]{MIRO loss with regularization term for feature extractor-2.}
            $X_{clean|B|}^{weak}$ = weak-augmentation($B_{clean}$)\\
            $X_{noisy|B|}^{weak}$ = weak-augmentation($B_{noisy}$)\\
            $X_{clean|B|}^{strong}$ = strong-augmentation($B_{clean}$)\\
            $X_{noisy|B|}^{strong}$ = strong-augmentation($B_{noisy}$)\\
            Get labeled set with UNICON label refinement on clean batch.\\
            Get unlabeled set with UNICON pseudo label on noisy batch.\\
            $L_{u1}, L_{u2}$ = MixMatch on labeled and unlabeled sets \tcp*[l]{UNICON unsupervised loss for feature extractor-1 and extractor-2.}
            Get $L_{c1}, L_{c2}$  \tcp*[l]{UNICON contrastive loss for feature extractor-1 and extractor-2.}
            Get $L_{r1}, L_{r2}$  \tcp*[l]{UNICON regularization loss for feature extractor-1 and extractor-2.}
            $loss_1 += \lambda_u* L_{u1} + \lambda_c* L_{c1} + \lambda_r* L_{r1}$ \tcp*[l]{Update UNICON loss for feature extractor-1.}
            $loss_2 += \lambda_u* L_{u2} + \lambda_c* L_{c2} + \lambda_r* L_{r2}$ \tcp*[l]{Update UNICON loss for feature extractor-2.}
            Update $f1_\theta$ and $f2_\theta$.
            Decide the start $step_s$ and end $step_e$ iteration for averaging in SWAD.
        }
    }
    $f1_{\theta'}$ = $\frac{1}{step_e - step_s+1} \Sigma f1_\theta$ 
    $f2_{\theta'}$ = $\frac{1}{step_e - step_s+1} \Sigma f2_\theta$ \tcp*[l]{SWAD parameter averaging.}
}
\end{algorithm}

\end{document}